\documentclass[runningheads]{llncs}
\usepackage{graphicx}

\usepackage{tikz}
\usepackage{comment}
\usepackage{amsmath,amssymb} 
\usepackage{color}

\usepackage{soul, color}           
\usepackage{amsmath,amssymb}
\usepackage{amsfonts}       
\usepackage{subcaption}
\usepackage{multirow}
\usepackage{makecell}
\usepackage[inline]{enumitem}
\usepackage{wrapfig}
\usepackage{pifont}
\newcommand{\header}[1]{\paragraph{\textbf{#1}}}
\usepackage{orcidlink}

\begin{document}
\pagestyle{headings}
\mainmatter
\def\ECCVSubNumber{5712}  

\title{Overcoming Shortcut Learning in a Target Domain by Generalizing Basic Visual Factors from a Source Domain} 

\titlerunning{Overcoming Shortcut Learning using a Source Domain}
%
\author{Piyapat Saranrittichai\inst{1, 2}\orcidlink{0000-0003-0620-7945} \and
Chaithanya Kumar Mummadi \inst{1, 2}\orcidlink{0000-0002-1173-2720}\index{Mummadi, Chaithanya Kumar} \and
Claudia Blaiotta \inst{1}\orcidlink{0000-0003-2314-3939} \and
Mauricio Munoz \inst{1}\orcidlink{0000-0002-9520-4430} \and
Volker Fischer\inst{1}\orcidlink{0000-0001-5437-4030}}
\authorrunning{P. Saranrittichai et al.}
%
\institute{Bosch Center for Artificial Intelligence \and University of Freiburg}
\maketitle

\begin{abstract}

Shortcut learning occurs when a deep neural network overly relies on spurious correlations in the training dataset in order to solve downstream tasks. Prior works have shown how this impairs the compositional generalization capability of deep learning models. To address this problem, we propose a novel approach to mitigate shortcut learning in uncontrolled target domains. Our approach extends the training set with an additional dataset (the source domain), which is specifically designed to facilitate learning independent representations of basic visual factors. We benchmark our idea on synthetic target domains where we explicitly control shortcut opportunities as well as real-world target domains. Furthermore, we analyze the effect of different specifications of the source domain and the network architecture on compositional generalization. Our main finding is that leveraging data from a source domain is an effective way to mitigate shortcut learning. By promoting independence across different factors of variation in the learned representations, networks can learn to consider only predictive factors and ignore potential shortcut factors during inference.

\end{abstract}


\section{Introduction}

Humans seamlessly categorize objects in the real world by their basic visual factors (e.g., shape, texture, color). For example, we perceive a red fire truck as a object with \emph{red} color and the shape of a \emph{fire truck}. We are able to do this because we have learned abstract concepts of shape and color, which easily generalize to common objects including unseen, out-of-distribution (OOD) data \cite{zeithamova2020generalization}. Unlike humans, modern deep neural networks (DNNs) do not possess a generalized notion of basic visual factors and therefore tend to perform poorly on OOD data. This is especially true when networks exploit shortcuts inherent in the data, i.e., when they excessively rely on visual factors that are easy to learn and predictive on in-distribution data but fail to generalize on OOD samples \cite{geirhos2020shortcut,xiao2020noise}. For example, DNNs trained on a dataset in which all fire trucks are red might use color as a shortcut to recognize fire trucks while ignoring the more semantically meaningful attribute of shape. As a result, such a network may misclassify a yellow fire truck as a school bus.

In this work, we study the ability of DNNs to recognize OOD samples in the context of compositional generalization, i.e., the ability to understand unseen combinations of known elements. 
While a few approaches have already been proposed to improve generalization in compositional zero-shot recognition tasks \cite{nagarajan2018attributes,atzmon2020causal,purushwalkam2019task}, the performance of all these methods heavily depends on the number of combinations observed during training. This is due to the fact that a low number of seen combinations generates more shortcut opportunities \cite{geirhos2020shortcut}. A na\"ive remedy to shortcut learning is to collect missing combinations but this could be costly as some combinations can be rare in practice. In this work, we aim to investigate the underpinning of compositional generalization in DNNs and hence focus on the challenging and unexplored regime in which different visual attributes are fully-correlated in the training data (e.g. where one attribute such as color is a deterministic function of another one, such as shape). In the rest of this paper, we will refer to the domain in which we aim to achieve compositional generalization as the \emph{target domain}.

\begin{figure}[t]
    \centering
    \includegraphics[width=0.6\textwidth]{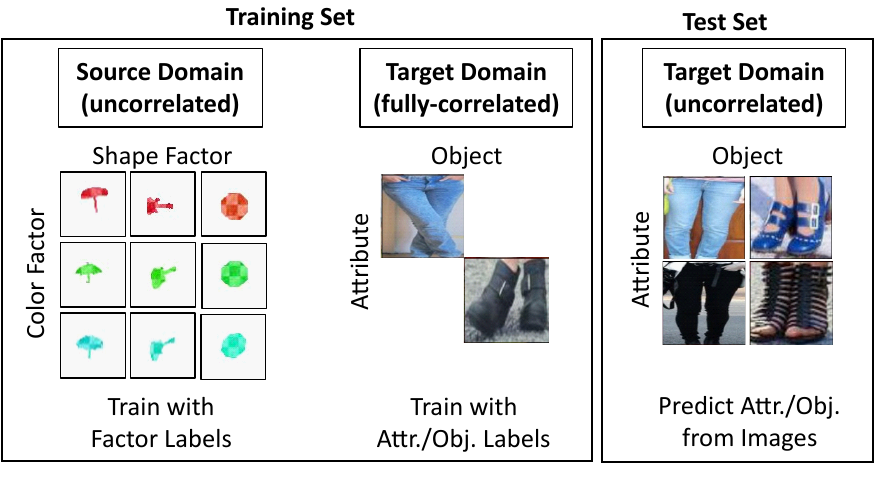}
    \caption{Our task is to predict attribute/object labels in the target domain whose correlated training labels induce high shortcut opportunities. To improve generalization, we propose to augment the training set with a cheap source domain to learn representations of visual factors. In this given example, attribute and object in the target domain associate to color and shape factors respectively.}
    \label{fig:teaser_setup}
\end{figure}

We propose a novel approach to mitigate shortcut learning in such a target domain by introducing a well-controlled \emph{source domain} in the form of an additional training dataset, which can be cheaply generated. This source dataset is specifically designed to facilitate learning of basic visual factors by constraining them to be uncorrelated (all factor combinations appear uniformly during training). With this source dataset, we aim to learn independent representations of generic visual factors in order to improve OOD generalization in the target domain. Overview of our setup is shown in Figure \ref{fig:teaser_setup}. We show practical benefits of such a simple and easy-to-generate source domain for improving the performance across multiple, more complex, target domains. Our approach provides a low-cost and effective strategy to improve compositional generalization.

Considering our source dataset, we employ DiagVib \cite{eulig2021diagvib}, a framework to generate datasets whose visual factors (i.e., shape, color, texture, lightness and background) can be customized (see Figure \ref{fig:source_dataset}). These factors are suitable for our purpose for the following reasons. Firstly, they are generic for common objects. Additionally,
as suggested by previous studies \cite{geirhos2018imagenet,xiao2020noise}, some visual factors (e.g., background and texture) are likely to introduce spurious correlations while other factors (e.g., shape) are more robust when used for recognizing objects. By including these key factors in the source dataset, our models can learn to focus more on important factors, ignore potential shortcut factors and thus improve the models’ robustness to shortcuts.

Our contributions are as follows:
firstly, we introduce a novel framework to improve compositional generalization in fully-correlated target domains. This is achieved by leveraging a source domain in order to alleviate shortcut learning. Secondly, we propose a simple network architecture exploiting the source domain to learn independent representations of visual factors. We also show that, if the target domain is not strictly fully-correlated, we can require less \emph{a priori} knowledge for our approach.
Lastly, we perform ablation studies to investigate effects of different source dataset configurations.
Our main finding is that the source domain can act as a regularizer that encourages the internal representations of basic visual factors in DNNs to be less entangled. We show that this consistently improves compositional generalization across different target domains.

\section{Related Works}

\header{Compositional Generalization (CG)}
We study CG in the context of compositional zero-shot learning, where the goal is to recognize images of unseen attribute-object combinations given only some combinations seen during training. A simple baseline VisProd \cite{misra2017red,nagarajan2018attributes} uses multiple classifiers to predict attribute and object labels. More recently, \cite{nagarajan2018attributes,purushwalkam2019task,atzmon2020causal,li2021learning,naeem2021learning,mancini2021open} learn to map images and labels (i.e., attribute-object combinations) to their joint feature space.
It is commonly assumed that the same object can be seen in combination with certain number of attribute values during training. In our work, we study the corner case where the number of seen combinations is minimal, e.g., a fully-correlated attribute-object combinations setting, which introduces severe shortcut opportunities. A recent work \cite{atzmon2020causal} shows dramatic performance degradation of DNNs when reducing the number of seen combinations. The objective of this work is to mitigate this effect by incorporating a suitable source domain.
We remark that, in most CG approaches, the label space has only two dimensions (i.e., attribute and object types). Instead, in our work the label space of the source domain is not restricted to two dimensions but can be as high-dimensional as the number of annotated factors. In particular, we consider basic visual factors as considered in \cite{eulig2021diagvib}, which are more likely to generalize across different tasks.


\header{Domain Generalization}
The aim of domain generalization is to learn models that generalize to unseen domains at test time. The problem we address in this work is therefore related to this line of research.
More specifically, our problem setting loosely resembles \textit{Heterogeneous Domain Generalization} (HeDG) 
\cite{zhou2020deep,zhou2020learning,yiying2019feature,li2019episodic,wang2020heterogeneous}, 
because we assume that the label distributions in the source and target domains have different, possibly disjoint, support (e.g., the labels represent objects in the source domain and animals in the target domain).
Nevertheless, it should be noted that our work stems from a different motivation compared to the domain generalization literature. 
Unlike domain generalization, we do not assume the target data distribution to be unavailable during training. Instead, we assume to have access to a heavily biased sample from the target domain. 
From this perspective, our setup is also related to domain adaptation \cite{ganin2015unsupervised,chang2019domain}, 
where training is performed on both the source and target domains together.

\header{Learning Independent Representations}
The topic of compositional generalization is also linked to the notion of disentanglement in generative modeling. The goal of disentangled representation learning is to construct a compact and interpretable latent representation, by discovering independent factors of variation (FoVs) in the data \cite{bengio2013representation,higgins2016beta,sauer2021counterfactual}.
Most methods proposed in the disentanglement literature assume statistical independence between the factors of variation and perform learning without supervision. Thus, they are predominantly trained and evaluated on synthetic data where the ground truth FoVs are perfectly uncorrelated \cite{locatello2019challenging}. This is an idealized setting, which is almost never encountered in the real world. Therefore, the usefulness and generalization ability of these methods when the training data is biased remains unclear. For instance, a recent large-scale empirical study found that several state-of-the-art methods from the disentanglement literature fail to disentangle pairs of correlated factors \cite{trauble2021disentangled}. 
Our work is also concerned with learning independent representations of basic visual factors, but, as opposed to prior works, we specifically focus on the problem of mitigating shortcut learning. It should be noted that, in contrast to unsupervised representation learning, we train representations of visual factors with factor annotations. However, we will discuss a scenario in which unsupervised representation learning can be integrated into our framework in section \ref{sec:experiments_learn_association}.

\section{Methodology}

\subsection{Problem Formulation}
\label{sec:method_problem}

Our task measures generalization performance in the presence of shortcuts from the perspective of compositional generalization. Specifically, let us consider a \emph{target domain/dataset} $t$, where each sample consists of an image $x_t \in \mathcal{X}_{t} \subset \mathcal{I}$ containing a single object. Such object is associated with one attribute: $y_t = (a, o) \in \mathcal{Y}_{t} = \mathcal{A}_{t} \times \mathcal{O}_{t}$, where $\mathcal{A}_{t} = \{a_1, a_2, \ldots\}$ and $\mathcal{O}_{t} = \{o_1, o_2, \ldots\}$ are sets of attribute and object type values respectively. In our task, not all attribute-object combinations are seen during training. The goal is to learn a model for the joint probability distribution $p(a, o\mid x_t)$ which yields good predictive performance on images of both seen and unseen attribute-object combinations.

The target dataset may be heavily biased, and thus naively training a classifier to predict object and attribute types can lead to shortcut learning resulting in poor OOD generalization. We define a dataset in a target domain, to be \emph{fully-correlated} when each annotated attribute label appears together with only one annotated object label or vice versa. We aim to investigate whether introducing an additional, specifically designed, dataset from a different \emph{source domain} $s$ can discourage DNNs from exploiting shortcuts when trained for attribute and object prediction in the target domain. For this purpose, each image $x_s \in \mathcal{X}_{s} \subset \mathcal{I}$ in the source domain $s$ is assumed to be labeled with a tuple $y_s = (f_1, f_2, \ldots, f_K) \in \mathcal{F}_s^{1} \times \mathcal{F}_s^{2} \times \ldots \times \mathcal{F}_s^{K}$, where $K$ is the number of factors and each $\mathcal{F}_s^{k}$ denotes the set possible values for an individual basic visual factor capturing generic image properties such as shape, color, texture, etc.

To utilize source factor information in the target domain, we build an association matrix that models the connection between different source factor representations to target attribute or object.
For fully-correlated target domains, we assume that the association is given as \emph{a priori} knowledge regarding the possible source factor(s) the model is expected to rely to avoid shortcuts. For example, in the Color-Fruit dataset (Figure \ref{fig:target_dataset_fruits_train}/\ref{fig:target_dataset_fruits_test}), attribute and object can be manually associated to color and shape factors respectively. This \emph{a priori} knowledge requirement can be relaxed in semi-correlated setting as presented in section \ref{sec:learning_factor_association}.

\subsection{Network Architecture}
\label{sec:method_network}

\begin{figure*}[t]
    \centering
    \begin{subfigure}[b]{0.45\textwidth}
        \centering
        \includegraphics[width=0.8\textwidth]{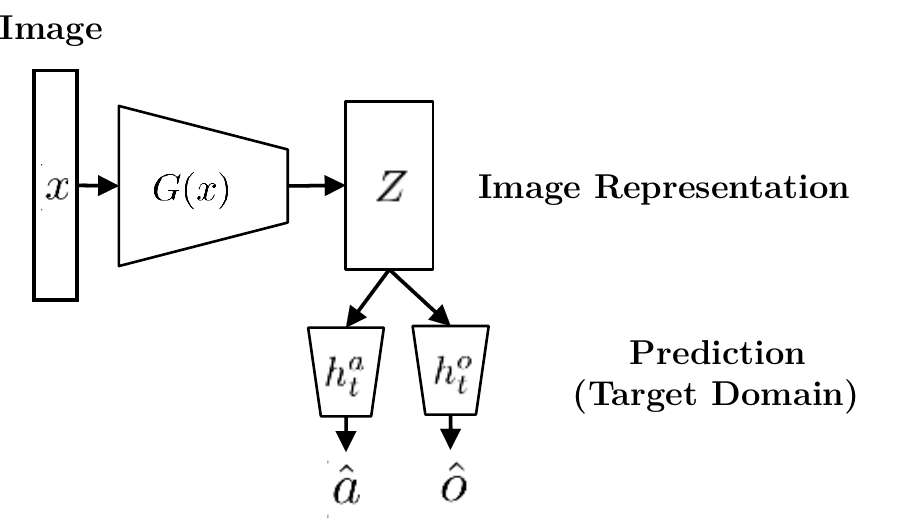}
        \caption{Global Image Representation}
        \label{fig:architecture_global}
    \end{subfigure}
    \begin{subfigure}[b]{0.45\textwidth}
        \centering
        \includegraphics[width=\textwidth]{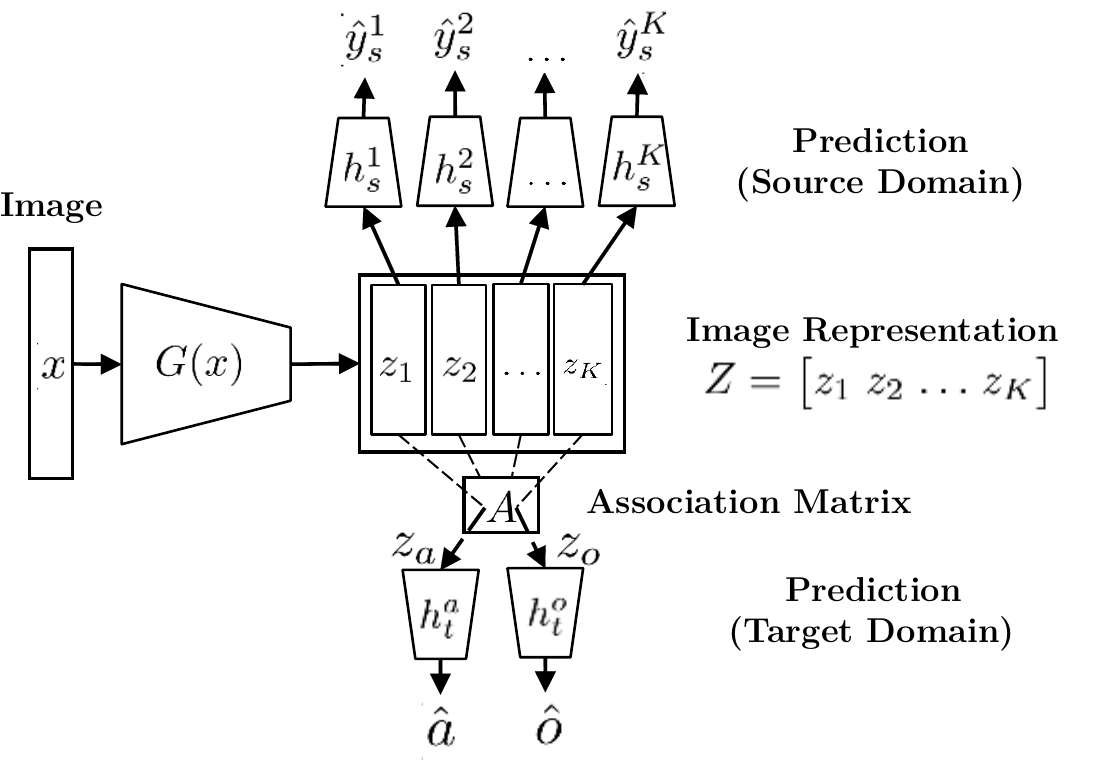}
        \caption{Factor Image Representation}
        \label{fig:architecture_factorized}
    \end{subfigure}
    \caption{Architectures with different image representations. a) Global: a single vector contains all visual clues. b) Factor: multiple factor representations encode different visual factors provided in source domain.}
    \label{fig:architecture}
\end{figure*}

\header{Architecture} 
We study a simple baseline model (similar to VisProdNN \cite{nagarajan2018attributes}), which naturally lends itself to exploiting the availability of a source dataset.
The baseline model (Figure \ref{fig:architecture_global}) consists of an encoder backbone $G$ and multiple prediction heads. It maps an input image to a global latent representation, which is then used to predict both the attribute and object labels. A crucial limitation of this model is that, without additional inductive biases, its latent representation will encode all predictive image factors. This makes the model more vulnerable to learning shortcuts, since it will be free to rely on easy to learn, predictive signals irrespective of their (in)ability to generalize to novel combinations \cite{NEURIPS2020_71e9c662}.

We extend the model of Figure \ref{fig:architecture_global} by splitting the latent representation into multiple factor representations, as in Figure \ref{fig:architecture_factorized}. As opposed to Figure \ref{fig:architecture_global}, the encoder $G$ produces $K$ non-overlapping factor representations. Each representation is intended to contain only information related to its corresponding basic factor (e.g., shape, color, texture). The encoder's output can then be written as $G(x) = Z = \begin{bmatrix} z_{1} & z_{2} & \ldots &z_{K} \end{bmatrix} \in \mathbb{R}^{D \times K}$ ($D$ is the size of a factor representation).

The prediction heads are divided in two subsets for predicting labels in source and target domains respectively. While in principle we could feed $Z$ as an input to all the prediction heads, as discussed above, this approach leads to poor compositional generalization. Instead, we introduce an additional inductive bias, namely that each source prediction logit should only depend on a single factor representation. Therefore, predictions for the source data $H_{s}$ are:
\begin{align}
\hat{y}_{s} = H_{s}(Z) = \begin{bmatrix} h_{s}^{1}(z_{1}) & h_{s}^{2}(z_{2}) & \ldots & h_{s}^{K}(z_{K}) \end{bmatrix} \;,
\end{align}
where $\hat{y}_{s} = \begin{bmatrix} \hat{y}_{s}^{1} & \hat{y}_{s}^{2} & \ldots & \hat{y}_{s}^{K} \end{bmatrix}$ contains the predicted factor values for a sample from the source domain. Ideally, each $\hat{y}_{s}^{k}$ should only depend on $z_{k}$ to discourage the latent representation from encoding information irrelevant to predict the $k$-th factor. Due to biases in the target domain, and in the absence of additional constraints, the architecture introduced above does not ensure invariance of every representation to the other factors. We explore different strategies to promote independence of the learned representations in section \ref{sec:constraints}.

In fully-correlated scenario, the association matrix can manually be defined as a binary matrix $A \in \left\{0, 1\right\}^{K \times 2}$ where $A_{k1}$ and  $A_{k2}$ are set to 1 only if the $k$-th factor informs the attribute and object prediction respectively. The representations of attribute ($z_a$) and object ($z_o$) can be obtained by $\begin{bmatrix} z_a & z_o \end{bmatrix} = ZA$. The prediction on target data can then be computed as follows:
\begin{align}
\hat{y}_{t} = H_{t}(ZA) =  \begin{bmatrix} h_{t}^{a}(z_a) & h_{t}^{o}(z_o) \end{bmatrix} \;,
\end{align}
where $\hat{y}_{t} = \left(\hat{a}, \hat{o} \right)$ is a tuple of the predicted attribute and object labels.

\header{Loss} To train the encoder and the predictors, we use a linear combination of the two loss terms $\mathcal{L}_{source} = \frac{1}{K} \sum_{\forall k} CE(\hat{y}_s^k, y_s^k)$ and $\mathcal{L}_{target} = \frac{1}{2}\sum_{l \in \left\{a, o\right\}} CE(\hat{y}_t^l, y_t^l)$, where $CE$ denotes the cross-entropy loss. $\lambda \geq 0$ is a hyperparameter weighting the importance of the regularizing loss term $\mathcal{L}_{source}$, which encourages a factor representation via the source samples. 

\header{Training} An  equal number of samples from the source and target domains are sampled for every minibatch and fed to the network in order to compute $\mathcal{L}_{source}$ and $\mathcal{L}_{target}$ separately. The network is optimized via gradient-based minimization of the total loss.
In this regard, all source samples will affect $G$ and $H_s$ and all target samples will affect $G$ and $H_t$.

\subsection{Additional Constraints}
\label{sec:constraints}

The factor representations $\{z_k\}_{k=1}^K$ may still be correlated when using the loss and model architecture described above. Consequently, in the target domain, $z_a$ (attribute representation) may be predictive of the object label, and $z_o$ (object representation) may be predictive of the attribute label. These are unintended shortcuts that lead to poor compositional generalization. We explore two additional constraints to further encourage independence among factor representations: the \emph{Isolated Latent Constraint} and the \emph{Cross Independence Constraint}.

\header{Isolated Latent (IL) Constraint} completely prevents factor representations from being influenced by the target domain. While this suppresses the effect of biases in the target dataset, it may harm discriminative performance in the target domain. This constraint is implemented by stopping gradients from $\mathcal{L}_{target}$ to the encoder $G$.

\header{Cross-Factor Independence (CI) Constraint} promotes independence of factor representations, by adding a set of small auxiliary networks $\left\{H'_{k_{1}k_{2}}\right\}_{k_{1} \neq k_{2}}$ for cross-factor predictions. While each $H'_{k_{1}k_{2}}$ is trained to predict $y_s^{k_{2}}$ from $z_{k_{1}}$, $G$ is trained to produce $Z$ such that all $H'$ are poor predictors. More details are presented in the appendix \ref{sec:appendix_ci_source}. Although the CI constraint only encourages independence with respect to the source domain, we investigate if this property can be transferred to target domains in section \ref{sec:experiments_freezed_vs_xpred}.

\subsection{Learning Factor Association Matrix $A$}
\label{sec:learning_factor_association}

In non-fully-correlated target domains (e.g., semi-correlated setting mentioned in section \ref{sec:experiments_learn_association}), we can relax the requirement that the association matrix $A$ must be manually given \emph{a priori} and, instead, learn it in an end-to-end fashion together with the network parameters. To this end, we apply a continuous relaxation of the binary matrix $A$, and allow it to contain real numbers within $[0, 1]$ such that each column sums to 1 (i.e. overall weightage across source factors) to maintain scales of the attribute/object representations using the softmax function.

We found that na\"ively learning $A$ without any additional constraints can lead to poor properties of the association matrix. Ideally, the association matrix should match the target attribute or object type to only one (or a few) robust source factor(s), and ignore factors vulnerable to shortcuts. For this reason, we propose to add an additional regularization $\mathcal{L}_{Reg} = \alpha\mathcal{L}_{Entropy} + \beta\mathcal{L}_{Suppress}$.

$\mathcal{L}_{Entropy}$ is the sum of the entropy values of both columns of $A$. Minimizing this entropy loss will reduce the number of source factors used to predict target properties, encouraging only robust factors to be considered when minimizing together with cross-entropy losses. On the other hand, $\mathcal{L}_{Suppress}$ applies a regularization along the rows of $A$ to make sure no same source factor is predictive of both target predictions. In particular, for each row $i$, if its maximum value $A_{ij^{max}}$ is higher than a threshold $\tau$, all other entries will be suppressed by adding $\left(\sum_{j \neq j^{max}}{A_{ij}}\right)*\left(\text{sg}(A_{ij^{max}}) - \tau\right)$ to the loss term. The symbol $\text{sg}(A_{ij^{max}})$ indicates the stop gradient operation, such that minimizing $\mathcal{L}_{Suppress}$ only affects the cell whose values are not maximum in each row. Detailed experiments on the automatic learning of the association matrix are presented in section \ref{sec:experiments_learn_association}.

\section{Experiments}
\label{sec:experiments}

We conduct several experiments to understand how incorporating the source domain affects compositional generalization in scenarios which are vulnerable to shortcut learning. We start by describing our experiment setup below.

\header{Datasets}

\begin{figure}[t]
  \centering
  \includegraphics[width=0.50\textwidth]{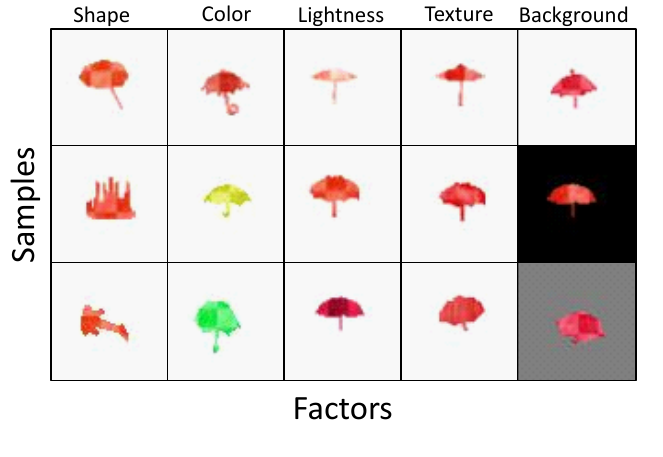}
  \caption{Illustration of DiagVib-Caltech (main source dataset) showing its multiple independent factors. More details of each factor is presented in the appendix \ref{sec:dv_dataset_configuration}.}
  \label{fig:source_dataset}
\end{figure}

\begin{figure*}[t]
    \centering
    \begin{subfigure}[b]{0.16\textwidth}
        \centering
        \includegraphics[width=\textwidth]{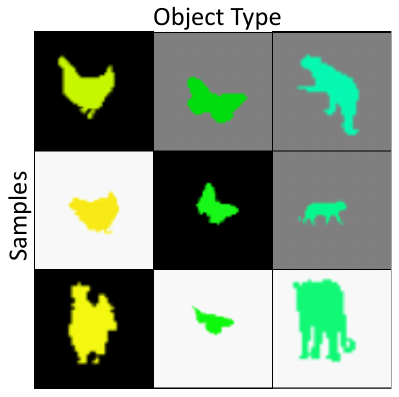}
        \caption{\makecell{DV-Animal \\ (Train)}}
        \label{fig:target_dataset_animal_train}
    \end{subfigure}
    \begin{subfigure}[b]{0.16\textwidth}
        \centering
        \includegraphics[width=\textwidth]{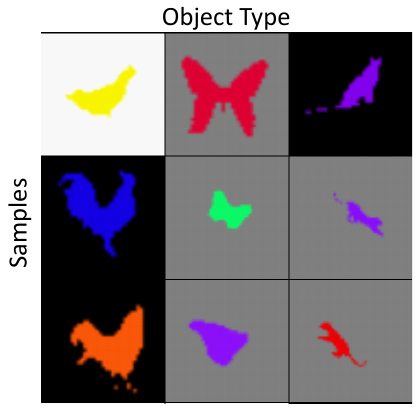}
        \caption{\makecell{DV-Animal \\ (Test)}}
        \label{fig:target_dataset_animal_test}
    \end{subfigure}
    \begin{subfigure}[b]{0.16\textwidth}
        \centering
        \includegraphics[width=\textwidth]{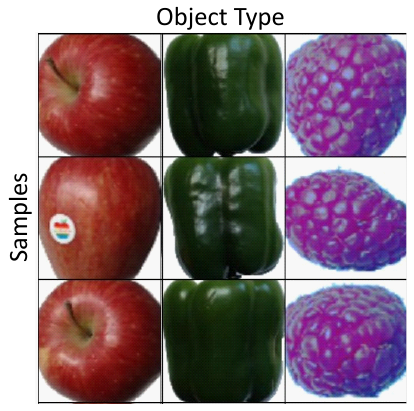}
        \caption{\makecell{Color-Fruit \\ (Train)}}
        \label{fig:target_dataset_fruits_train}
    \end{subfigure}
    \begin{subfigure}[b]{0.16\textwidth}
        \centering
        \includegraphics[width=\textwidth]{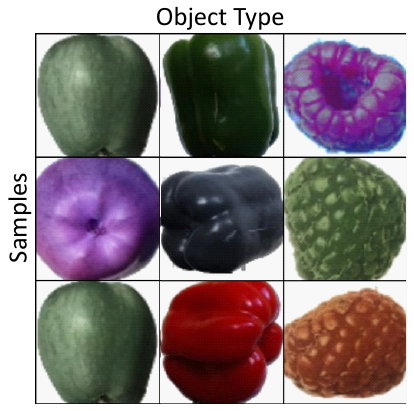}
        \caption{\makecell{Color-Fruit \\ (Test)}}
        \label{fig:target_dataset_fruits_test}
    \end{subfigure}
    \begin{subfigure}[b]{0.16\textwidth}
        \centering
        \includegraphics[width=\textwidth]{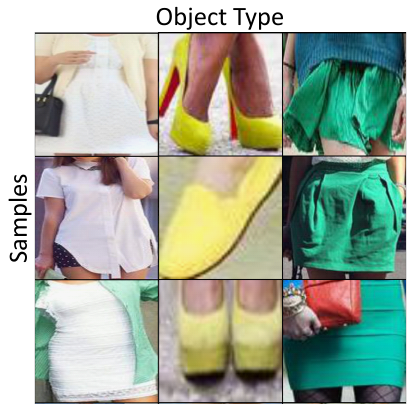}
        \caption{\makecell{Color-Fashion \\ (Train)}}
        \label{fig:target_dataset_fashion_train}
    \end{subfigure}
    \begin{subfigure}[b]{0.16\textwidth}
        \centering
        \includegraphics[width=\textwidth]{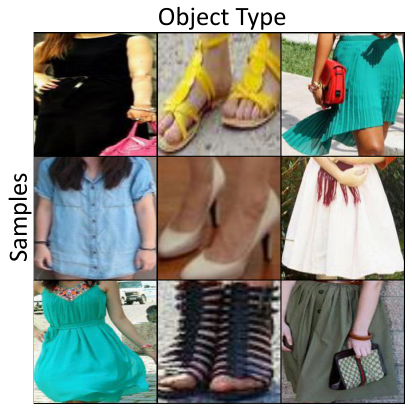}
        \caption{\makecell{Color-Fashion \\ (Test)}}
        \label{fig:target_dataset_fashion_test}
    \end{subfigure}    
    \caption{Samples from target domains. A column of each grid corresponds to an object type. Notice that each object (shape) corresponds to a single attribute (color) during training (see \subref{fig:target_dataset_animal_train}, \subref{fig:target_dataset_fruits_train} and \subref{fig:target_dataset_fashion_train}) but not during testing (see \subref{fig:target_dataset_animal_test}, \subref{fig:target_dataset_fruits_test} and \subref{fig:target_dataset_fashion_test}).}
    \label{fig:target_dataset}
\end{figure*}

We use the following datasets in our experiments:

\emph{DiagVib} \cite{eulig2021diagvib} We extend the original framework to use shapes other than MNIST to increase variances.
Our main configurations based on this framework are
\begin {enumerate*} [label=\itshape\alph*\upshape)]
	\item \emph{DiagVib-Caltech}: With 50 non-animal shapes from Caltech101 \cite{marlin2010inductive} in 12 colors. 5 basic visual factors are available as shown in Figure \ref{fig:source_dataset}.  This dataset is our main source dataset. We will later show that compositional generalization can be improved by this low-cost dataset even in more complex target domains.
	\item \emph{DiagVib-Animal}: With 10 animal shapes from \cite{bai2009integrating} in 10 colors on 3 backgrounds and scales (see Figure \ref{fig:target_dataset_animal_train}/\ref{fig:target_dataset_animal_test}).
\end {enumerate*}

\emph{Color-Fruit} This dataset is comprised of real fruit images (of 5 types) from the Fruit-360 dataset \cite{murecsan2017fruit}. Additionally, we control colors of the fruits using the recolorization approach from \cite{zhang2017real} (see Figure \ref{fig:target_dataset_fruits_train}/\ref{fig:target_dataset_fruits_test}). More details on this dataset generation are described in the appendix \ref{sec:appendix_fruit_generation}.

\emph{AO-CLEVR} This dataset is proposed in \cite{atzmon2020causal} to benchmark compositional generalization. It contains 3 basic shapes in 7 different colors. We will use this dataset as an additional target domain. We simulate correlation between attributes and objects by limiting one color to appear with only one shape during training.

\emph{Color-Fashion} This dataset is originally proposed in \cite{liu2013fashion} in which each image sample depicts a person dressed with cloth combinations. For each sample, cloth type and color segmentations are provided. In this paper, original images are cropped so that only one cloth type is appeared in individual images (see Figure \ref{fig:target_dataset_fashion_train}/\ref{fig:target_dataset_fashion_test}). 5 cloth types (T-shirt, skirt, jeans, shoes and dress) and 5 colors (Black, White, Yellow, Green and Blue) are selected from the original dataset.

For all target domains in our work, if a manual factor association matrix $A$ is required for any algorithms, attribute and object types (i.e., shapes, fruit types, cloth types) are associated to color and shape factors respectively. This is intuitive since shape can robustly predict object types in general. Poorly assigned association can lead to poor performance as detailed in the appendix \ref{sec:appendix_association_matrix_importance}.

\header{Evaluation}
We evaluate compositional generalization in the open world setting as in \cite{purushwalkam2019task} by computing predictive accuracies on seen and unseen attribute-object combinations as well as their harmonic mean (HM). One difference to \cite{purushwalkam2019task} is that, for fair comparison, we do not use an additional hyperparameter (i.e. bias term) to calibrate the likelihood of predicting unseen attribute-object combinations (More details regarding the bias term are described in the appendix \ref{sec:appendix_evaluation_with_bias}). During testing, any attribute-object combinations in the target domain can appear. All quantitative results are averaged across 6 different random training seeds.

\header{Network Configurations}
We follow the same convention as in \cite{nagarajan2018attributes,atzmon2020causal,purushwalkam2019task} by using ImageNet pretrained features from ResNet-50 as network inputs. $G$, $H_s$ and $H_t$ are modeled as fully-connected networks. Factor-0 and FactorSRC refers to our architecture when a source domain is ignored and used, respectively. IL and CI constraints are appended as suffixes if the constraints are applied. Also, we benchmark the architecture with the global image representation (Figure  \ref{fig:architecture_global}). Similarly, Global-0 and GlobalSRC denote its configurations where a source domain is not incorporated and incorporated respectively. More implementation details can be found in the appendix \ref{sec:appendix_implementation}. The code for our implementation is publicly available at \url{https://github.com/boschresearch/sourcegen}.

\header{Baselines} We compare our approach against the following baselines (1) \textbf{LabelEmbed+} \cite{nagarajan2018attributes}: a vanilla baseline, which performs recognition with a joint feature space for images and labels. (2) \textbf{TMN} \cite{purushwalkam2019task}, which employs automatic network rewiring conditioned on attribute-object pair hypotheses. (3) \textbf{CGE} \cite{naeem2021learning}, which exploits graph structure to regularize the joint feature space. 

\subsection{Compositional Generalization in Fully-Correlated Scenario}
\label{sec:source_generalization}

We investigate the impact of using an uncorrelated source domain (all factor combinations can appear uniformly) across different target domains whose labels are fully-correlated. We compare several variants of our approach against baselines that do not use a source domain. The results are shown in Table \ref{table:performance_overview}.

We begin by noting that, without a source domain (first five rows), baselines generally perform well only on the seen combinations but not on the unseen ones. E.g., Factor-0 on the Color-Fruit dataset has seen accuracy of $100\%$ compared to only $2.9\%$ on unseen combinations (Low seen accuracies of some baselines are discussed in the appendix \ref{sec:appendix_evaluation_with_bias}). This occurs as these networks are trained only on the target datasets with correlated combinations so that they learn to excessively exploit the easiest predictive visual factors present in the datasets. On the unseen combinations, these predictive factors do not necessarily generalize to the intended labels. This degrades the generalization of the networks.

In contrast, using a source domain (see FactorSRC variations) consistently improves the HM accuracy by increasing accuracy on unseen combinations, at the expense of a partial loss of performance on seen combinations. For example, the FactorSRC-IL baseline on the Color-Fruit dataset has a seen accuracy of $95.5\%$, which is lower compared to baselines that do not use the source domain, but in turn exhibits the highest unseen accuracy ($40.7\%$). This shows a reduction of shortcut learning. The same trend holds for all other datasets we consider, with different seen/unseen accuracy trade-offs across datasets.

Compared to previous works, our results show that a \emph{single} and \emph{simple} source domain improves generalization performance on unseen combinations across different target domains. An alternative \emph{na\"ive} solution would be to collect data corresponding to unseen combinations directly in the target domain and include them in the training set. However, this is in practice not a viable solution: not only is collecting data in a real-world target domain expensive but, perhaps more importantly, the biases that affect the training set are often unknown, which makes collecting an uncorrelated dataset difficult in practice. Rather, generalization improvement on those target domains can be achieved without much additional costs if we have a \emph{universal} (a source domain can be used for multiple target domains) source domain at hand that can be \emph{generated cheaply}. One open problem is the trade-off between generalization and in-distribution accuracy, which in some cases is still sub-optimal, especially when the target domain has a large domain shift from the source domain (see the drop of seen accuracy of the FactorSRC-IL on the Color-Fashion dataset). Improving this trade-off is an open research question leaving a scope for improvement in future works.

\header{Global vs Factor Image Representations}
Our results suggest that a factor representation is essential to exploit the source domain for compositional generalization. In fact, all GlobalSRC variations, in spite of incorporating the source domain during training, exhibit significantly lower unseen accuracy compared to the FactorSRC variants. This is due to the fact that the global representation contains information about all visual factors that are relevant for the object-attribute prediction task, thus offering easy-to-learn shortcuts that harm generalization performance. In contrast, by using factor representations, we promote factor disentanglement such that shortcuts are harder to learn.

\begin{table*}[t]
\caption{Accuracies on DiagVib-Animal, Color-Fruit, AO-CLEVR (each has random chance accuracy of 1\%, 4\% and 4.7\% respectively) and Color-Fashion target domains. DiagVib-Caltech is used as the source domain.}
\centering
\fontsize{7}{11}\selectfont
\begin{tabular}{cc|ccc|ccc|ccc|ccc}
\Xhline{4\arrayrulewidth}
\multirow{2}{*}{Approach} & \multirow{2}{*}{\makecell{Use \\ Source?}} & \multicolumn{3}{c|}{DiagVib-Animal} & \multicolumn{3}{c|}{Color-Fruit} & \multicolumn{3}{c|}{AO-CLEVR} & \multicolumn{3}{c}{Color-Fashion} \\ 
 && Seen & Unseen & HM & Seen & Unseen & HM & Seen & Unseen & HM & Seen & Unseen & HM \\ \hline
LabelEmbed+ & \ding{55} & $69.6$ & $7.3$ & $13.2$ & \textbf{100} & $6.8$ & $12.5$ & \textbf{100} & $0.7$ & $1.5$ & $37.0$ & 7.1 & 11.9 \\
TMN & \ding{55} & 95.5 & $0.1$ & $0.3$ & \textbf{100} & $0.0$ & $0.0$ & \textbf{100} & $0.0$ & $0.0$ & $86.5$ & 0.7 & 1.4 \\
CGE & \ding{55} & 43.8 & 9.1 & 15.0 & 84.4 & 7.8 & 14.3 & 94.0 & 9.3 & 16.9 & 21.6 & \textbf{20.5} & 21.0 \\
Global-0 & \ding{55} & \textbf{96.1} & $0.0$ & $0.1$ & \textbf{100} & $1.5$ & $3.0$ & \textbf{100} & $0.3$ & $0.5$ & \textbf{93.6} & 0.0 & 0.0 \\
Factor-0 & \ding{55} & $95.2$ & $0.8$ & $1.5$ & \textbf{100} & $2.9$ & $5.5$ & \textbf{100} & $2.2$ & $4.3$ & $92.7$ & 1.8 & 3.6 \\ \hline
GlobalSRC & \ding{51} & $94.2$ & $0.3$ & $0.5$ & \textbf{100} & $1.1$ & $2.2$ & \textbf{100} & $0.3$ & $0.7$ & $85.5$ & 0.2 & 0.4 \\
GlobalSRC-IL & \ding{51} & 92.4 & 0.3 & 0.7 & \textbf{100} & 0.7 & 1.4 & 98.9 & 0.8 & 1.6 & $61.8$ & 2.2 & 4.2 \\
FactorSRC & \ding{51} & $90.0$ & $7.0$ & $13.0$ & $99.7$ & $27.3$ & $42.4$ & $99.9$ & $3.2$ & $6.3$ & $76.4$ & 8.3 & 15.0 \\
FactorSRC-CI & \ding{51} & $91.2$ & $7.9$ & $14.5$ & \textbf{100} & $10.9$ & $19.6$ & \textbf{100} & $2.3$ & $4.5$ & $87.3$ & 8.2 & 14.8 \\
FactorSRC-IL & \ding{51} & $56.3$ & \textbf{32.6} & \textbf{41.3} & $95.5$ & \textbf{40.7} & \textbf{57.0} & $89.5$ & \textbf{19.6} & \textbf{32.1} & 32.7 & 17.0 & \textbf{22.3} \\ \Xhline{4\arrayrulewidth}
\end{tabular}
\label{table:performance_overview}
\end{table*}


\subsection{Impact of Additional Constraints}

\label{sec:experiments_freezed_vs_xpred}

\begin{table}[t]
\caption{Cross prediction accuracies on DiagVib-Animal. These are obtained by using each associated factor representation ($z_a$ or $z_o$) to predict each target label (attribute or object) with a linear model. Ideal independence representations will have high direct prediction accuracies (predict their own labels well) but low cross prediction accuracies (predict others' labels poorly).}
\centering
\fontsize{9}{11}\selectfont
\begin{tabular}{c|cc|cc}
\Xhline{4\arrayrulewidth}
\multirow{2}{*}{Approach} & \multicolumn{2}{c|}{Direct-Prediction $\uparrow$} & \multicolumn{2}{c}{Cross-Prediction $\downarrow$} \\
 & $z_a \rightarrow \hat{a}$ & $z_o \rightarrow \hat{o}$ & $z_a \rightarrow  \hat{o}$ & $z_o \rightarrow \hat{a}$ \\ \hline
FactorSRC & \textbf{62} & \textbf{86} & 77 & 44 \\
FactorSRC-CI & 54 & 83 & 73 & 35 \\
FactorSRC-IL & 58 & 85 & \textbf{46} & \textbf{23} \\ \Xhline{4\arrayrulewidth}
\end{tabular}
\label{tab:independency_heatmap}
\end{table}


We investigate the role of IL and CI constraints. We begin by noting that FactorSRC-IL gives the best HM accuracies across all target domains. Our hypothesis is that this result can be explained by a larger cross-factor information flow into the learned factor representations when adding the CI constraint compared to IL. To quantitatively measure the magnitude of cross-factor leakage, we extract $z_k$ from all test samples and use them to predict all labels ($y_s^1, y_s^2, \ldots y_s^K$ for a source domain or $\hat{a}, \hat{o}$ for a target domain) with linear models. Ideally, $z_k$ should predict its associated label (direct prediction) well but should fail to predict other labels (cross-prediction). We present these results in Table \ref{tab:independency_heatmap}.

First, we are interested whether the CI constraint can encourage independence, not only in the source domain, but more importantly in the target domain. We observe that, while independence is promoted by FactorSRC-CI in the source domain well (see the appendix \ref{sec:appendix_ci_source}), in the target domain, cross-prediction accuracies decrease only slightly compared to FactorSRC. Thereby, we can infer that the independence enforced by $\mathcal{L}_{H'}$ in the source domain is not necessarily transferable to the target domain. One possible reason is that factor representations are still affected by the dataset biases in the target domain via $\mathcal{L}_{target}$.

On the other hand, dataset bias in the target domain cannot affect the factor representations with the IL constraint. In Table \ref{tab:independency_heatmap}, although an explicit independent constraint is not introduced, the factor representations are less entangled in the target domain, which is indicated by lower cross-prediction accuracies (while high direct-prediction accuracies are preserved). The independence which is indirectly encouraged only by $\mathcal{L}_{source}$ enables shortcut-robust factor representations resulting in 
better HM accuracies in Table \ref{table:performance_overview}.

\subsection{Learning Association Matrix for Semi-Correlated Scenario}
\label{sec:experiments_learn_association}
\begin{table*}[t]
\caption{Accuracies on various target domains in semi-correlated scenarios.}
\centering
\begin{tabular}{c|ccc|ccc|ccc}
\Xhline{4\arrayrulewidth}
\multirow{2}{*}{Approach} & \multicolumn{3}{c|}{DiagVib-Animal} & \multicolumn{3}{c|}{Color-Fruit} & \multicolumn{3}{c}{Color-Fashion} \\ 
 & Seen & Unseen & HM & Seen & Unseen & HM & Seen & Unseen & HM \\ \hline
TMN & \textbf{83.4} & 3.2 & 6.2 & \textbf{99.9} & 3.1 & 6.0 & \textbf{68.6} & 3.7 & 6.9 \\
CGE & 51.8 & 1.9 & 3.6 & 72.0 & 8.4 & 15.0 & 42.5 & 13.0 & 19.9 \\
FactorSRC-IL & 52.8 & \textbf{35.0} & \textbf{42.1} & 93.7 & \textbf{36.3} & \textbf{52.3} & 32.5 & \textbf{15.0} & \textbf{20.5} \\
FactorSRC-IL-LA & 72.7 & 3.7 & 7.1 & 99.0 & 17.5 & 29.6 & 53.1 & 8.0 & 13.9 \\
FactorSRC-IL-LA\textsuperscript{R} & 60.9 & 24.2 & 34.7 & 96.0 & 32.7 & 48.6 & 35.2 & 14.3 & 20.2 \\
\Xhline{4\arrayrulewidth}
\end{tabular}
\label{table:semi_correlated_comparison}
\end{table*}


The association matrix $A$ must be known \emph{a priori} for fully-correlated target domains because the target data alone does not contain enough information to distinguish object types from their attributes. 
For a \emph{semi-correlated} target domain, a more general solution is viable, in which the association matrix is learned. In this case, each object type is observed in combination with at least two attribute values. The additional combinations make it possible to distinguish attribute from object type. In order to learn the association matrix $A$, we adopt the algorithm introduced in section \ref{sec:learning_factor_association}.

Table \ref{table:semi_correlated_comparison} reports performance of different algorithms. FactorSRC-IL-LA and FactorSRC-IL-LA\textsuperscript{R} indicate algorithms that learn the association matrix without and with association regularization respectively. Results suggest that na\"ively backpropagating through the matrix $A$ is not an effective strategy as indicated by the large gap between HM accuracy of FactorSRC-IL and FactorSRC-IL-LA. This is due to an undesired property of the association matrix which we will later investigate. Fortunately, this undesired property can be alleviated by incorporating our proposed regularization constraints during training, as indicate by the comparable performance of FactorSRC-IL and FactorSRC-IL-LA\textsuperscript{R}.

\begin{figure*}[t]
    \centering
    \begin{subfigure}[b]{0.32\textwidth}
        \centering
        \includegraphics[width=\textwidth]{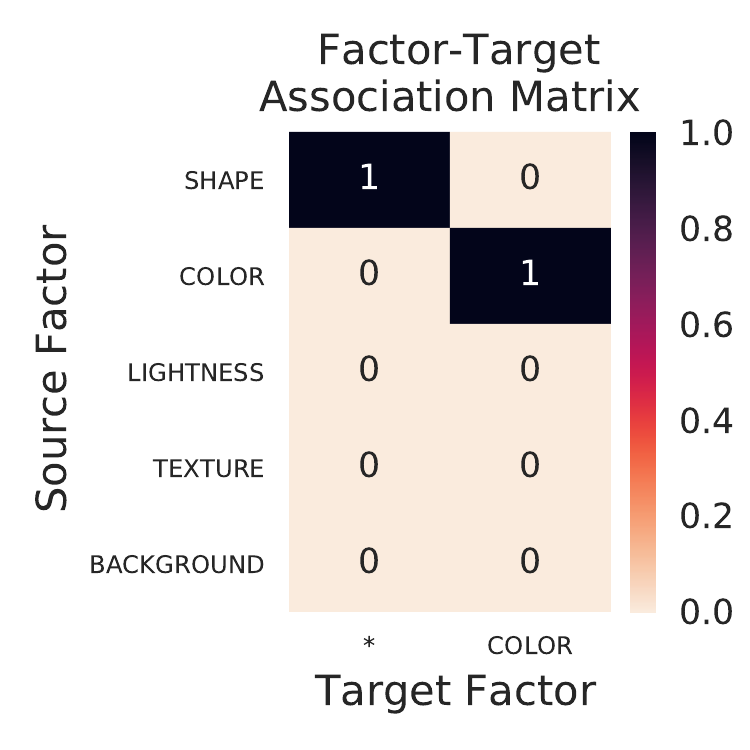}
        \caption{\makecell{Manual Association \\ Matrix}}
        \label{fig:learn_association_matrix_manual}
    \end{subfigure}    
    \begin{subfigure}[b]{0.32\textwidth}
        \centering
        \includegraphics[width=\textwidth]{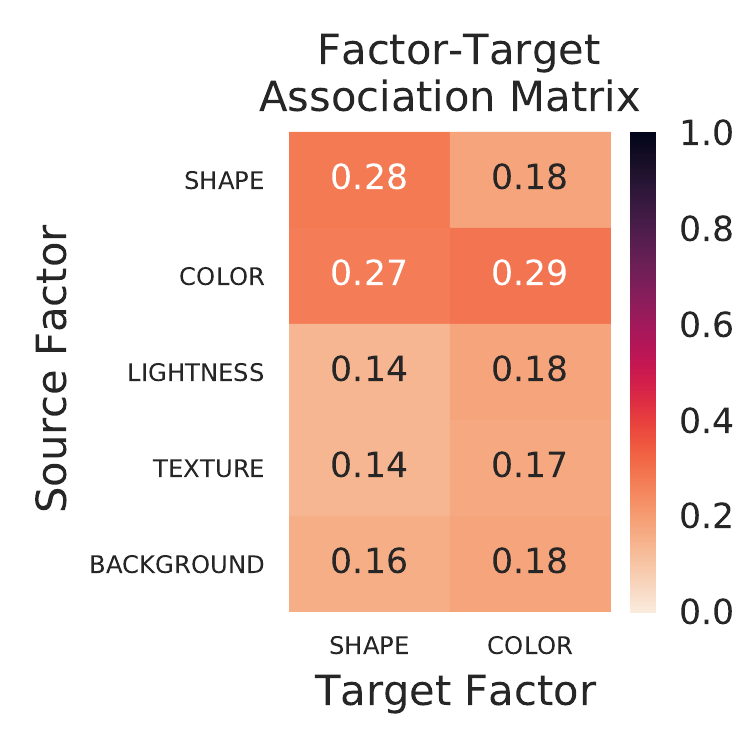}
        \caption{\makecell{DiagVib-Animal \\ (FactorSRC-IL-LA)}}
        \label{fig:learn_association_matrix_dv_la}
    \end{subfigure}
        \centering
    \begin{subfigure}[b]{0.33\textwidth}
        \centering
        \includegraphics[width=\textwidth]{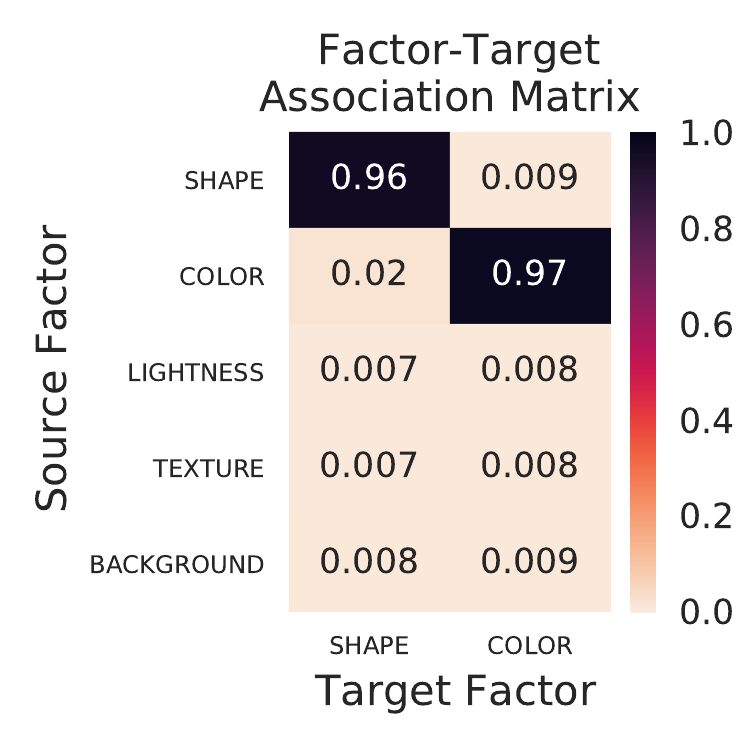}
        \caption{\makecell{DiagVib-Animal \\ (FactorSRC-IL-LA\textsuperscript{R})}}
        \label{fig:learn_association_matrix_dv_la_r}
    \end{subfigure}
    
    \begin{subfigure}[b]{0.32\textwidth}
        \centering
        \includegraphics[width=\textwidth]{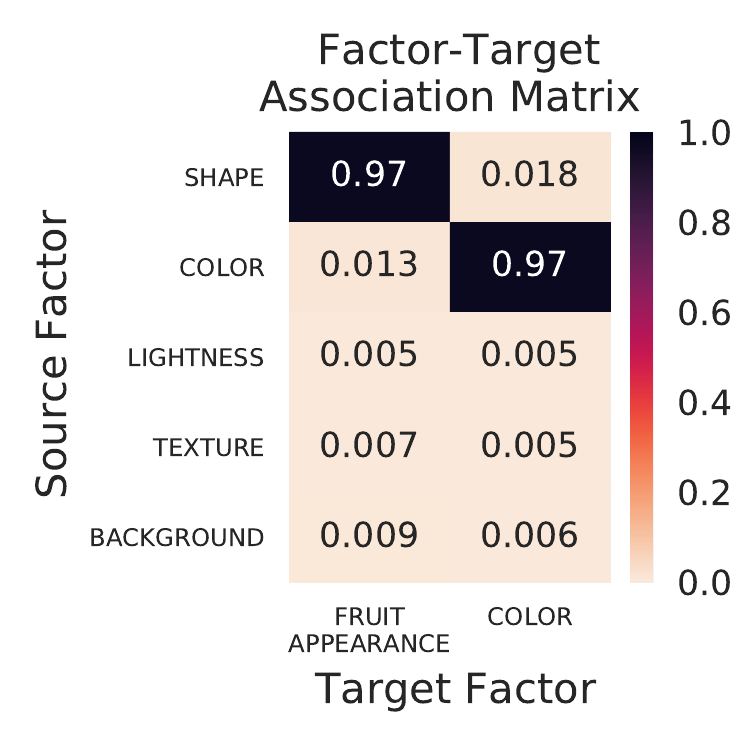}
        \caption{\makecell{Color-Fruit \\ (FactorSRC-IL-LA\textsuperscript{R} [1])}}
        \label{fig:learn_association_matrix_color_fruit_la_r_0}
    \end{subfigure}
    \begin{subfigure}[b]{0.32\textwidth}
        \centering
        \includegraphics[width=\textwidth]{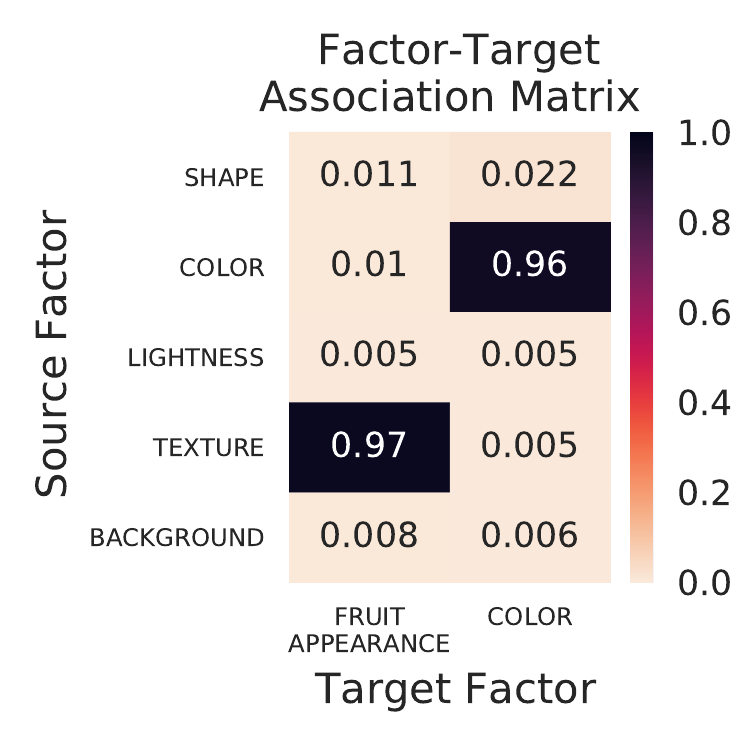}
        \caption{\makecell{Color-Fruit \\ (FactorSRC-IL-LA\textsuperscript{R} [2])}}
        \label{fig:learn_association_matrix_color_fruit_la_r_1}
    \end{subfigure}
    \begin{subfigure}[b]{0.32\textwidth}
        \centering
        \includegraphics[width=\textwidth]{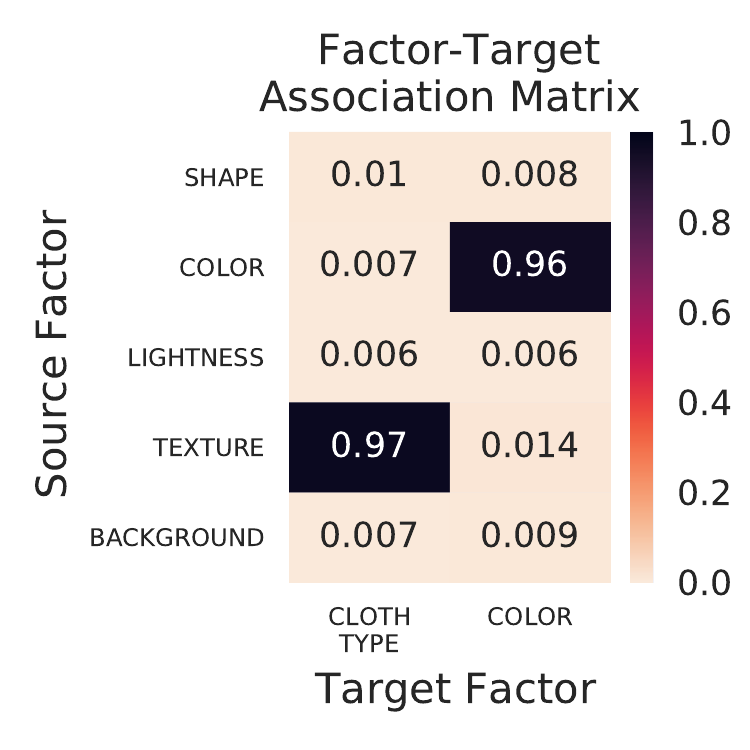}
        \caption{\makecell{Color-Fashion \\ (FactorSRC-IL-LA\textsuperscript{R})}}
        \label{fig:learn_association_matrix_color_fashion_la_r}
    \end{subfigure}    
    \caption{Factor association matrices learned from different target datasets with various approaches resulted from representative runs. Manual Association Matrix in Figure \subref{fig:learn_association_matrix_manual} is only used for our manual association (not the groundtruth).}
    \label{fig:learn_association_matrix}
\end{figure*}

To qualitatively assess the correctness of learned associations, we visualize $A$ as heatmaps in Figure \ref{fig:learn_association_matrix}. In addition, we compare the association matrix that we manually assigned in FactorSRC* (Figure \ref{fig:learn_association_matrix_manual}) and the learned ones. First we consider the simplest target dataset, DiagVib-Animal. In this case, FactorSRC-IL-LA, which does not apply any regularization, fails to retrieve the association matrix (Figure \ref{fig:learn_association_matrix_dv_la}). Even if the association matrix has the high weights that associate shape factor to shape target and color factor to color target correctly, the matrix is still far from sparse, making the model vulnerable to shortcuts (see results in Table \ref{table:semi_correlated_comparison}). On the other hand, when regularization constraints are introduced in FactorSRC-IL-LA\textsuperscript{R}, the learned matrix is more sparse and closer to the manually assigned matrix (see Figure \ref{fig:learn_association_matrix_dv_la_r}), which results in lower shortcut vulnerability of the model and higher compositional generalization performance.

In the more realistic case of the Color-Fruit dataset, we find that with multiple random seeds, FactorSRC-IL-LA\textsuperscript{R} can converge to two possible configurations for the estimated matrix $A$. The first configuration (Figure \ref{fig:learn_association_matrix_color_fruit_la_r_0}) is close to our manual association as it matches the fruit type to the shape factor. Another configuration, on the other hand, associates the fruit type to the texture factor (Figure \ref{fig:learn_association_matrix_color_fruit_la_r_1}). This is not surprising, since in the color-fruit dataset, both shape and texture factors are predictive of fruit type. The same observation has already been made in the context of conventional image classification \cite{geirhos2018imagenet} especially with complex object shapes. We observed the same behaviour in the case of the Color-Fashion dataset where the learned matrix associates garment type to the texture factor (Figure \ref{fig:learn_association_matrix_color_fashion_la_r}). Another advantage of our approach indicated by this observation is that it yields higher model interpretability,  as we can understand which visual factors are important for network predictions.

The fact that learning of the association is possible when target datasets are not fully-correlated makes our approach well suited to practical applications. As the association is determined automatically, minimal \emph{a priori} knowledge is required. In other words, factor representations can also be learned from images by any approaches including unsupervised representation learning that disentangle image representations into multiple independent factor representations \cite{higgins2016beta}.

\subsection{Properties of the Source Domains}

We also investigate the properties of source domains which encourage generalization. Our main findings can be summarized as follows: first, basic visual factors represented in the source domain should be sufficiently diverse and aligned with visual properties of the target data. Second, the fact that all factor combinations are available in the source domain during training is crucial. This allows deep networks to learn meaningful representation for each factor. Lastly, large intra-class variation of factors is also important to encourage better generalization. More details of our ablation studies can be found in the appendix \ref{sec:dv_dataset_configuration}.
	
\section{Conclusion}

We study vulnerability of DNNs to shortcuts by evaluating their compositional generalization on target domains with correlated attribute-object combinations.
We provide empirical evidence that incorporating an additional source domain can improve generalization on unseen combinations on target domains.
The source domain enables certain networks to represent inputs in terms of multiple independent visual factors. From our findings, the impact of the source domain on compositional generalization relies on two major conditions: (1) Choice of network model: networks should have internal representations in which visual factors are disentangled and independent with respect to target domains (2) Choice of source domain: the source domain should be uncorrelated and cover main basic factors.
The fact that our source domain is simple also shows a practical benefit that performance on certain target tasks can efficiently improve using an easy-to-generate source domain. This is relatively cheaper compared to acquiring samples from complex target domains. If target domains are not fully-correlated, some requirements of manual labels/annotations can be relaxed, leading to more practical applications of this work.
We hope this work will serve as an inspiration to integrate inductive biases in the forms of datasets and network design.



\clearpage
%
%

\clearpage
\appendix
\appendix

\section{Appendix}

In this section, we present arrays of ablation studies to understand crucial properties of the source domains. All experiments in this section are performed in the fully-correlated target domains.

\subsection{DiagVib Dataset Configurations}
\label{sec:dv_dataset_configuration}

As mentioned in the experiments in section \ref{sec:experiments}, we use datasets based on the DiagVib framework which allows generation of synthetic datasets with custom configurations of basic visual factors. We consider five factors whose number of possible values are listed according to Table \ref{table:synthetic_dataset_spec}.  It should be noted that DiagVib-Caltech and DiagVib-Animal have different number of available shapes. 

\begin{table*}[ht!]
\caption{Different visual factors, which can be configured in the DiagVib framework}
\centering
\begin{tabular}{|c|l|r|}
\hline
\textbf{Factor} & \makecell{\textbf{Description}} & \makecell{\textbf{No. of Classes}} \\ \hline
Shape & Object boundary defined by a silhouette & \makecell[r]{Caltech: 50 \\ Animal: 10} \\ \hline
Color & Hue value in HSV space & 12 \\ \hline
Lightness & Lighting condition (e.g., bright, dark) & 4 \\ \hline
Texture & \makecell[l]{Pattern drawn inside the object \\ (e.g. wooden, checkerboard)} & 5 \\ \hline
Background & Background color & 3 \\ \hline
\end{tabular}
\label{table:synthetic_dataset_spec}
\end{table*}


\subsection{Ablation Studies on the Source Domain}
\label{sec:experiments_ablation_on_source}

\subsubsection{Impact of Uncorrelation of Factors}

\begin{table}[!ht]
\caption{Accuracies of FactorSRC-IL in the target domain (DiagVib-Animal) with variations of source domains to demonstrate the impact of their uncorrelated factors.}
\centering
\fontsize{9}{11}\selectfont
\begin{tabular}{c|c|c|c}
\Xhline{4\arrayrulewidth}
 Source Setting & Images from & \makecell{Correlated \\ Factors} & \makecell{Target \\ HM Acc.} \\ \hline
 Uncorrelated & DiagVib-Caltech & False & \textbf{33.5 $\pm$ 1.0} \\
 Correlated & DiagVib-Caltech & True & $2.5\pm0.7$ \\
 Target & DiagVib-Animal & True & $1.7\pm0.4$      \\ \Xhline{4\arrayrulewidth}

\end{tabular}
\label{table:uncorrelated_source_help}
\end{table}

In this study, we aim to investigate whether the improvement in generalization performance after incorporating the source domain stems from uncorrelating visual factors. We compare the following source dataset settings: \begin {enumerate*} [label=\itshape\alph*\upshape)]
	\item Uncorrelated: all factor combinations are available
	\item Correlated: shape and color factors are one-to-one correlated
	\item Target: use correlated data sampled from the target distribution (DiagVib-Animal) for training
\end {enumerate*}. For a fair comparison, the number of target-associated factors (shape/color) are reduced to 10 for Uncorrelated and Correlated settings, so as to match the Target setting. Results in Table \ref{table:uncorrelated_source_help} indicate that the Uncorrelated setting yields significantly higher accuracy compared to others. This empirically shows that this improvement in OOD generalization is indeed due to the uncorrelated nature of the source dataset and not just a mere result of the increased dataset size.

\subsubsection{Impact of Shape Variations}

We conduct another experiments to understand if the complexity of the shapes provided in the source domain affects accuracies in the target domain. We modify the DiagVib-Caltech source domain to use MNIST shapes and compare it to the original setting with Caltech shapes (we use 10 shapes in both cases to be comparable). Table \ref{table:shape_variations} shows that the setting with MNIST shapes has lower accuracies. We believe that this is due to the fact that MNIST has less intra-class shape variation compared to Caltech. For example, the shape of the number ones are not much different across different samples. This degrades the generality of the learned shape representation. This experiment suggests that a primary concern when constructing a source dataset should be intra-class variability of each factor.

\begin{table}[!ht]
\caption{Accuracies of FactorSRC-IL with variations of shape in the source domain.}
\centering
\fontsize{9}{11}\selectfont
\begin{tabular}{c|c|c|c}
\Xhline{4\arrayrulewidth}
Shape & \makecell{DiagVib-Animal \\ HM. Acc} & \makecell{Color-Fruit \\ HM. Acc} & \makecell{AO-CLEVR \\ HM. Acc} \\ \hline
Caltech & \textbf{33.5 $\pm$ 1.0} & \textbf{56.0 $\pm$ 2.7} & \textbf{36.4 $\pm$ 1.8} \\
MNIST & $31.9\pm0.7$ & $46.9\pm3.2$ & $29.0\pm1.4$ \\ \Xhline{4\arrayrulewidth}
\end{tabular}
\label{table:shape_variations}
\end{table}

\subsubsection{Impact of Available Factors}

In this experiment, we study the effects of varying the number of basic visual factors represented in the source domain. According to the result in Table \ref{table:ilatents_multiple_sources}, while we find that increasing the number visual factors yields better performance overall, for some factors, the effect on different target domains is different. For instance, with DiagVib-Animal as a target, including the background as a factor in the source domain improves performance significantly, due to the fact that the target domain has variable background colors. In contrast, this effect is not observed on Color-Fruit, whose images have a constant background. Instead, learning lightness and texture can improve generalization performance since these two factors have high variation in this target domain (DiagVib-Animal doesn't have variations of lightness and texture). We can infer from this result that the performance in the target domain tends to be better if the source domain captures basic factors which are represented in the target domain.

\begin{table}[ht!]
\caption{HM Accuracies from FactorSRC-IL approach on DiagVib-Caltech source domain with different presence of factors (S, C, L, T, B correspond to Shape Color, Lightness, Texture and Background respectively).}
\centering
\fontsize{9}{11}\selectfont
\begin{tabular}{c|c|c}
\Xhline{4\arrayrulewidth}
\multicolumn{1}{c|}{Factors} & DV.-Animal & Color-Fruit \\ \hline
S/C & $28.6\pm1.2$ & $49.5\pm5.0$ \\
S/C/L & $29.3\pm1.1$ & $53.2\pm3.5$ \\
S/C/L/T & $31.5\pm0.5$ & \textbf{57.7$\pm$3.3} \\
S/C/L/T/B & \textbf{41.3$\pm$1.6} & $57.0\pm4.7$ \\ \Xhline{4\arrayrulewidth}
\end{tabular}
\label{table:ilatents_multiple_sources}
\end{table}


In summary, we have performed an array of ablation studies to analyze properties of the source domain which encourage better generalization in target domains. Firstly, we showed that visual factors in the source domain should be uncorrelated. This facilitates disentanglement of visual factors' representations, which in turn leads to less shortcut vulnerability. Secondly, we demonstrated that intra-factor variability is crucial in order for deep networks to learn generalizable representations. Lastly, visual factors encoded in the source domain should cover as many predictive features in the target domain as possible. We believe that these three aspects are among the most important criteria, which should guide practitioners towards choosing better source domains for augmenting biased training datasets.

\subsubsection{Impact of Variations of the Number of Factor Classes}
\label{sec:appendix_vary_n_factors}

\begin{figure}[!ht]
    \centering
    \begin{subfigure}[b]{0.45\textwidth}
        \centering
        \includegraphics[width=\textwidth]{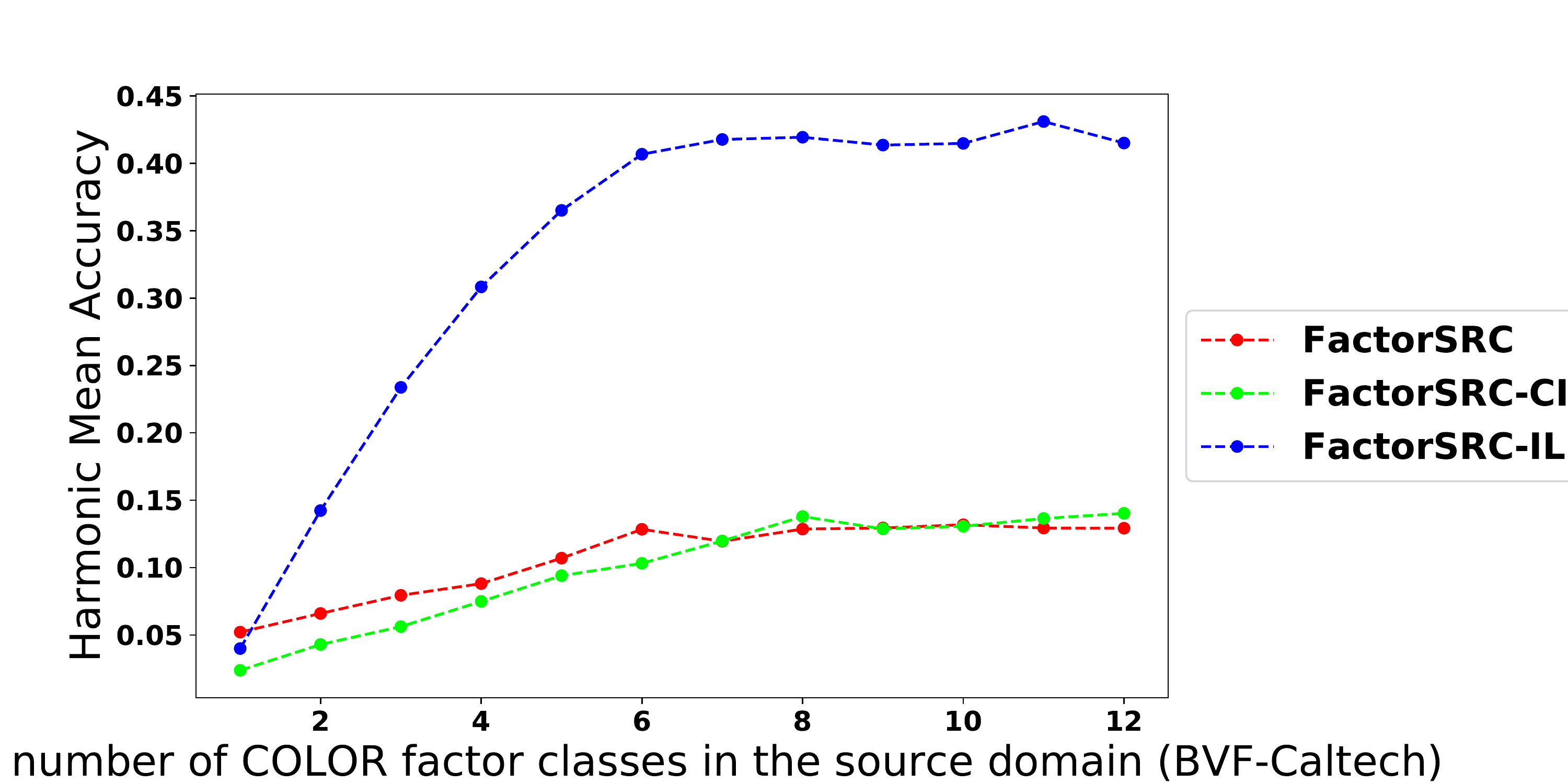}
        \caption{Varying the number of colors}
        \label{fig:vary_numbers_n_colors}
    \end{subfigure}

    \begin{subfigure}[b]{0.45\textwidth}
        \centering
        \includegraphics[width=\textwidth]{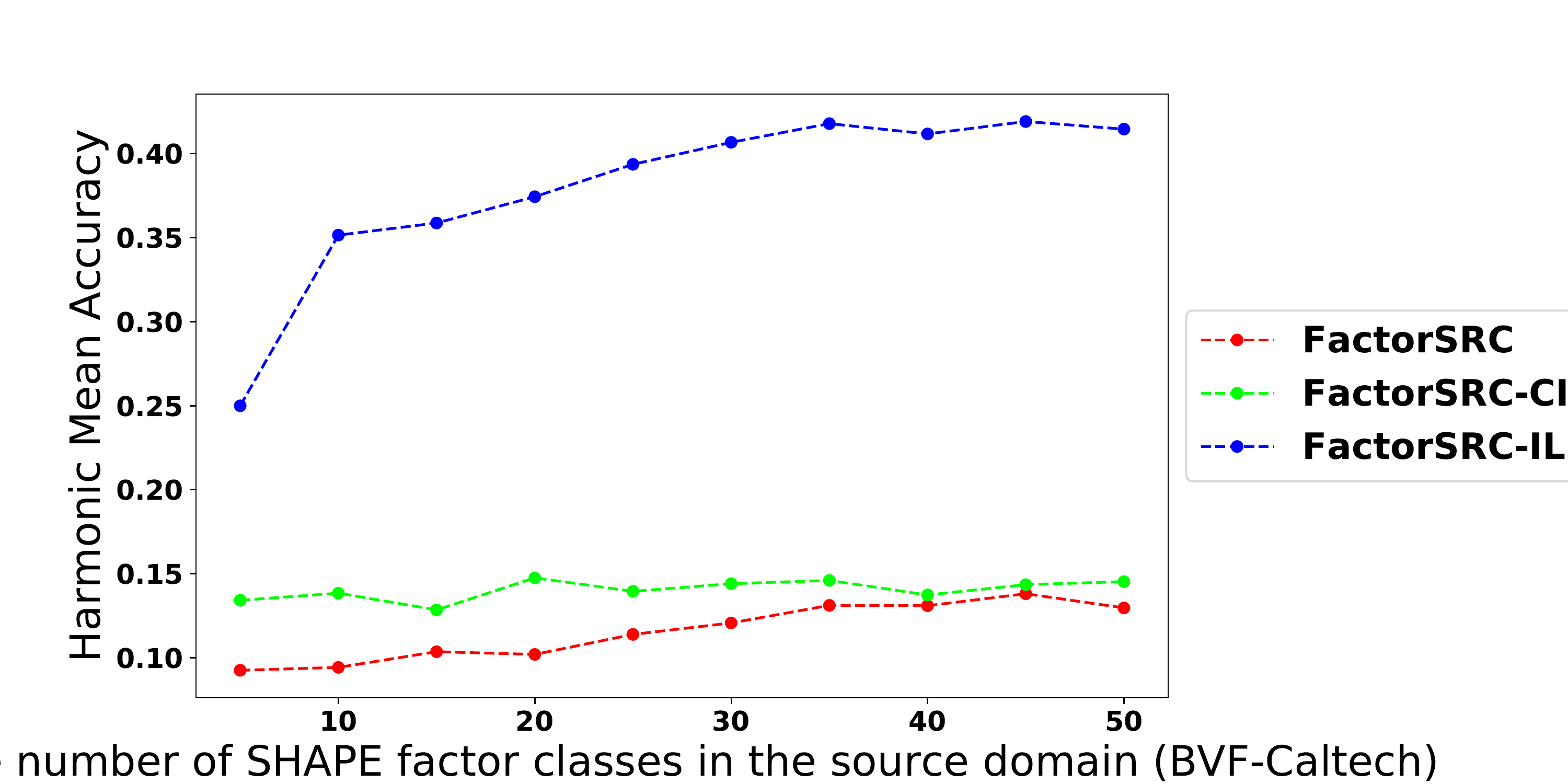}
        \caption{Varying the number of shapes}
        \label{fig:vary_numbers_n_shapes}
    \end{subfigure}
    \caption{Accuracies of FactorSRC-IL on the DiagVib-Animal with different number of factor classes (color and shape) while maintaining the same configuration for the other factors on the DiagVib-Caltech source domain.}
    \label{fig:vary_numbers}
\end{figure}

From our experiments, we showed that FactorSRC-IL can learn factor representations from the source domain, which improve compositional generalization in several target domains. In this section, we would like to investigate how the number of factor values in the source dataset affects generalization performance in the target domain. For this purpose, we vary the number of factor values associated to each target label (shape and color in our setting) and measure compositional generalization in the DiagVib-Animal target domain. Results are shown in Figure \ref{fig:vary_numbers} and indicate that a higher number of factor values generally leads to the better performance. This is intuitive since a higher number of classes should encourage networks to learn more general factor representations. An interesting observation is that the network needs only around 8 color classes to be close to optimal performance while around 35 classes are needed in the case of shape. We believe this is due to the fact that shape, as a basic visual factor, is more high-dimensional and thus more difficult to model than color. 

\subsection{Effect of the CI Constraint on the Source Domain}
\label{sec:appendix_ci_source}

\begin{figure*}
    \centering
    \begin{subfigure}[b]{0.45\textwidth}
        \centering
        \captionsetup{justification=centering}
        \includegraphics[width=\textwidth]{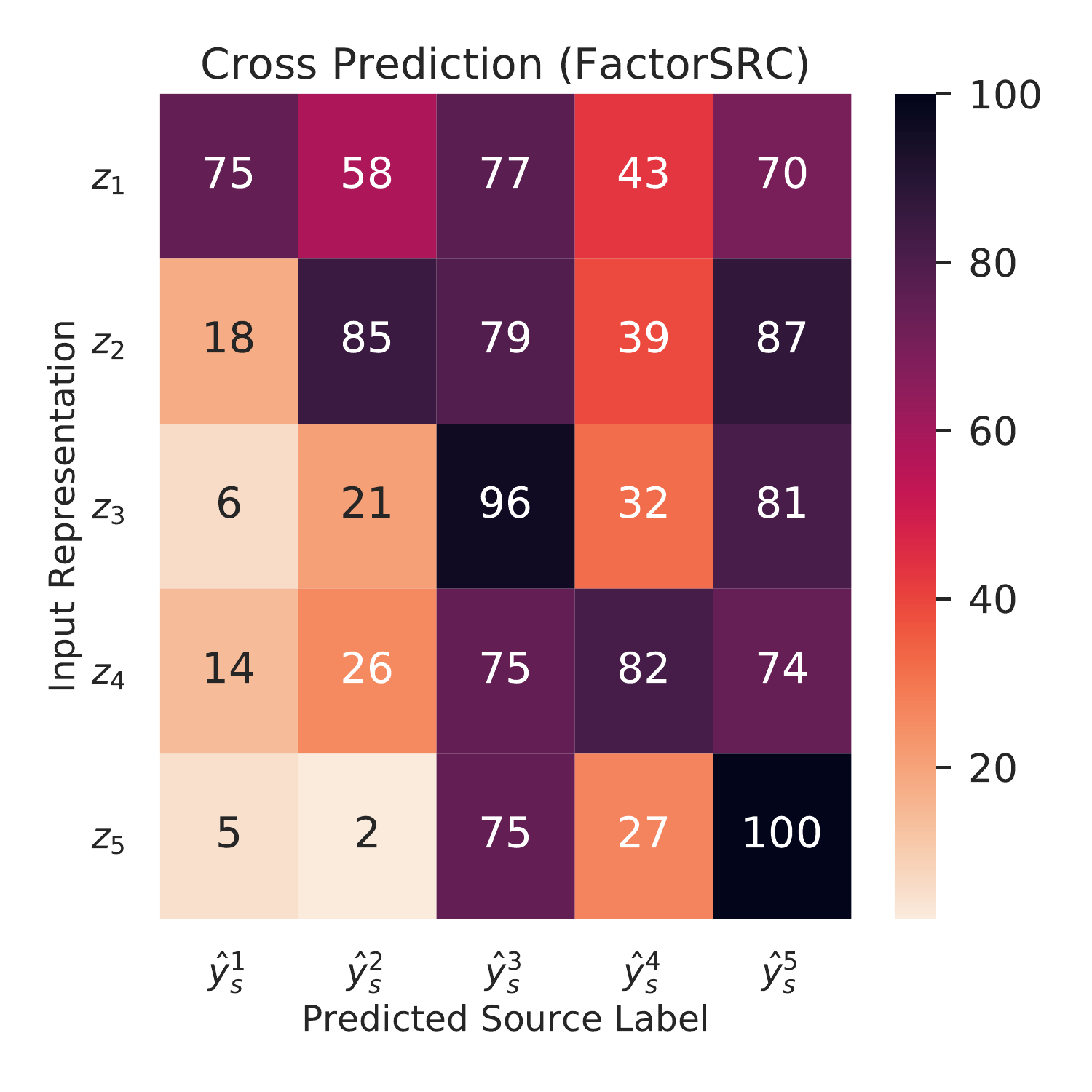}
        \caption{FactorSRC Heatmap}
        \label{fig:independency_heatmap_vanilla_source}
    \end{subfigure}
    \begin{subfigure}[b]{0.45\textwidth}
        \centering
        \captionsetup{justification=centering}
        \includegraphics[width=\textwidth]{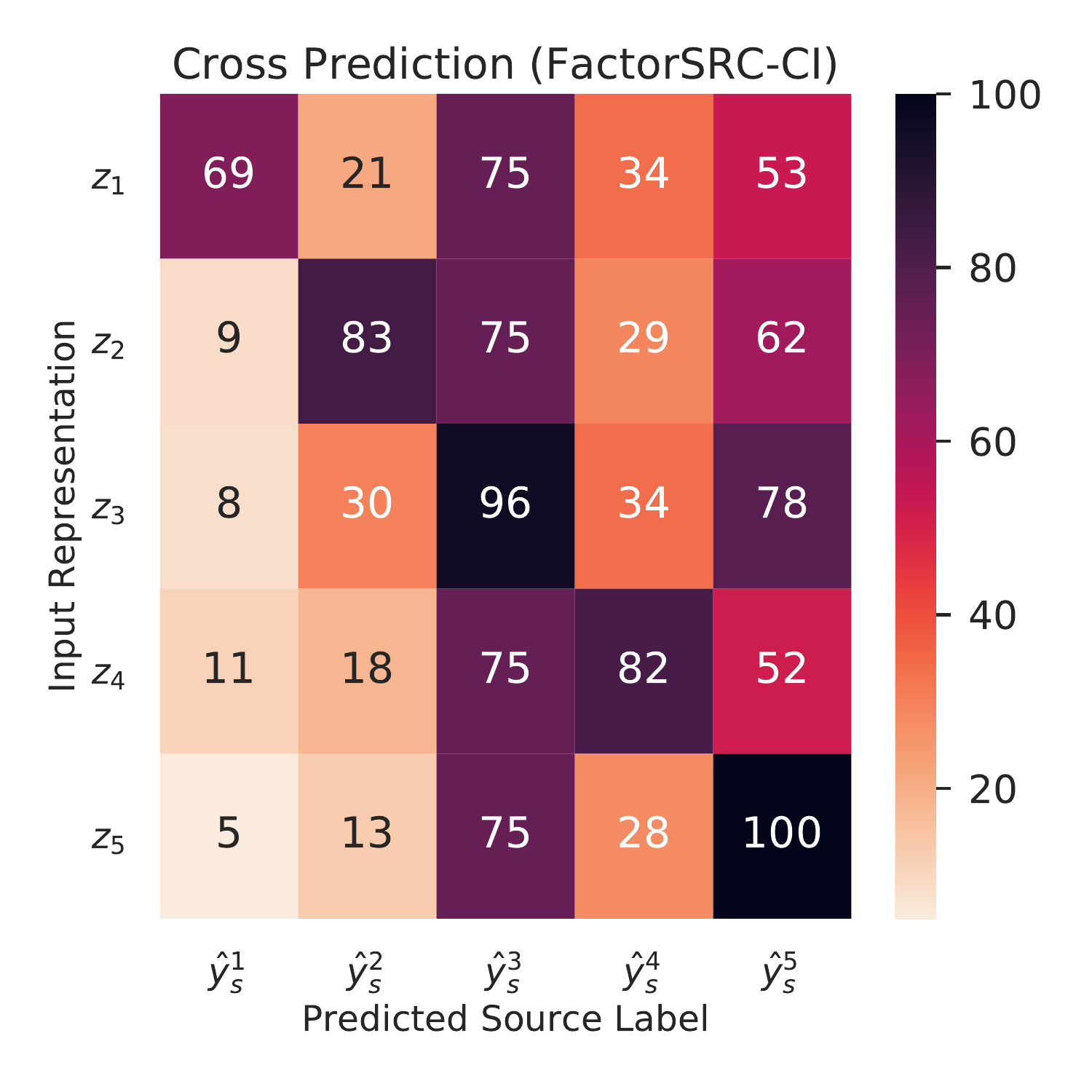}
        \caption{FactorSRC-CI Heatmap}
        \label{fig:independency_heatmap_xpred_source}
    \end{subfigure}
    \caption{Heatmaps displaying direct and cross-factor prediction accuracies using DiagVib-Caltech and DiagVib-Animal as source and target domain respectively. Each cell indicates the accuracy attained when a single factor representation ($z_k$ in each row) is used to predict labels ($\hat{y}_s^{l}$) with a linear model. A higher degree of independence among factor representations is expected to yield similar diagonal values but lower off-diagonal ones (cross-prediction). Factor indices from 1 to 5 correspond to shape, color, lightness, texture and background respectively.}
    
    \label{fig:independency_heatmap_source}
\end{figure*}

In our experiment section, we stated that the Cross-Factor Independence Constraint (CI) promotes independence of factor reresentations in the source domain. In this section, we provide experimental evidence supporting our claim. To this end, we compare cross-prediction accuracies with and without the CI constraint, for each factor among $z_1, z_2, \ldots, z_K$. Results are visualized in Figure \ref{fig:independency_heatmap_source}. We can see that, while direct-prediction accuracies are comparable with and without the CI constraint,  the cross-prediction performance decreases significantly when the CI constraint is introduced. This supports our hypothesis that the CI constraint induces a higher degree of independence among factor representations in the source domain.

\subsection{Importance of Association Matrix Assignment}
\label{sec:appendix_association_matrix_importance}

\begin{figure*}[ht!]
    \centering
    \begin{subfigure}[b]{0.48\textwidth}
        \centering
        \includegraphics[width=\textwidth]{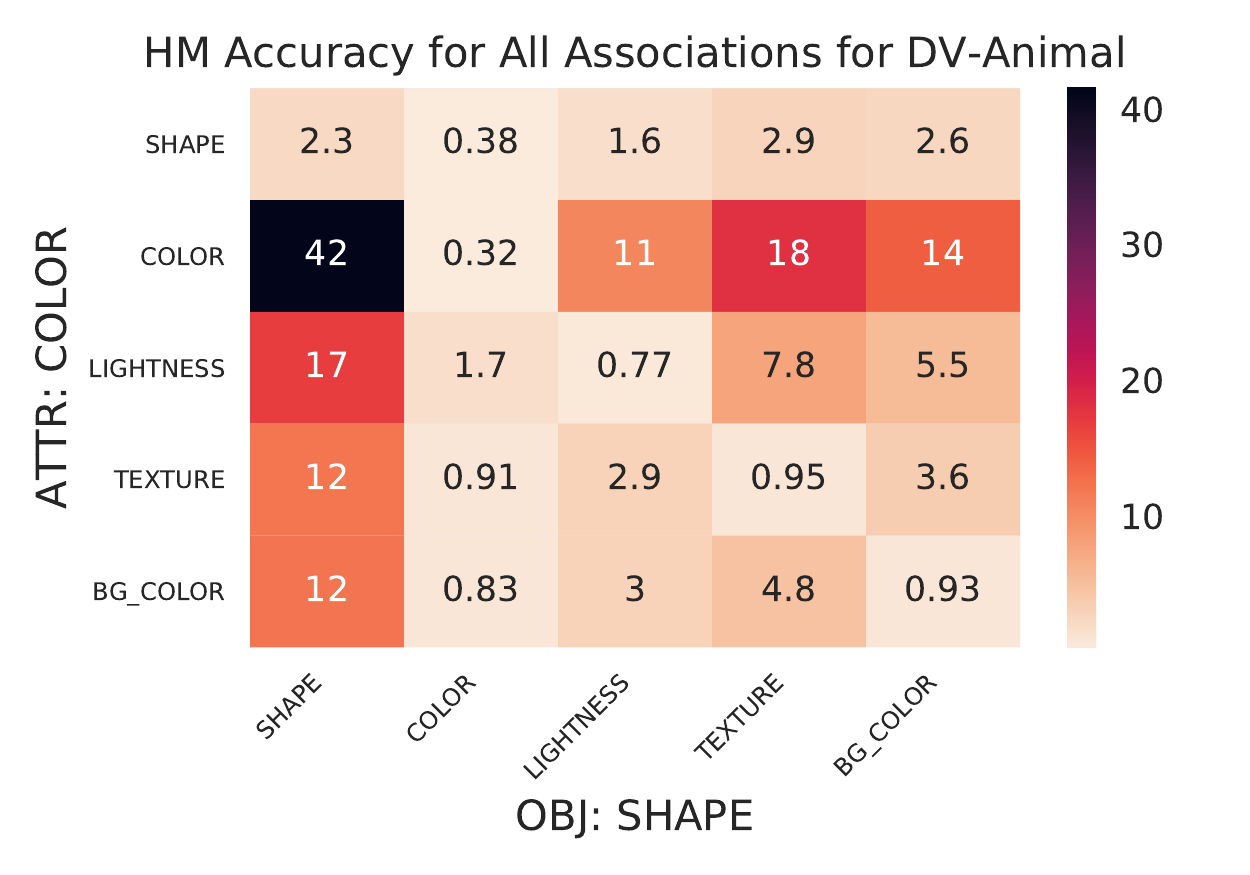}
        \caption{DiagVib-Animal}
        \label{fig:all_association_hm_dv}
    \end{subfigure}    
    \begin{subfigure}[b]{0.48\textwidth}
        \centering
        \includegraphics[width=\textwidth]{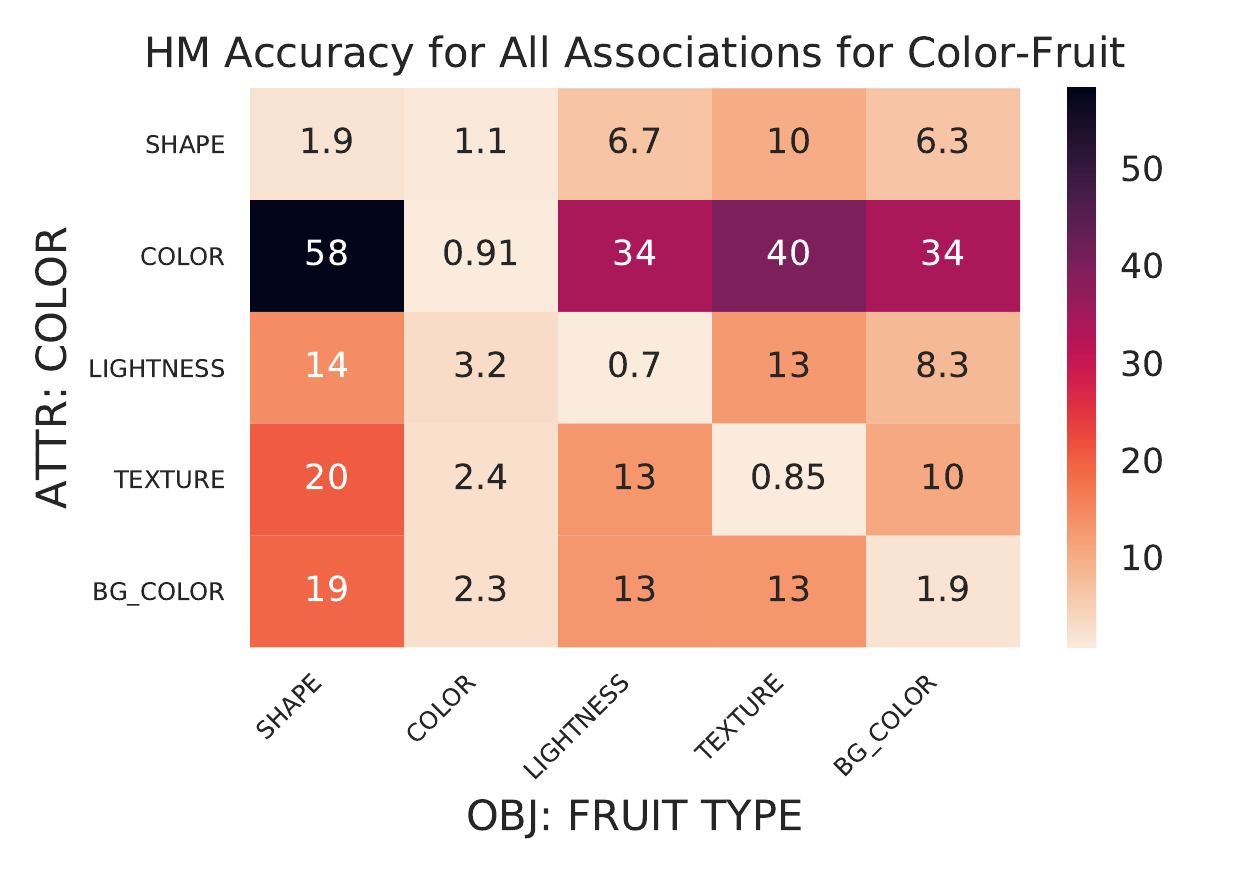}
        \caption{Color-Fruit}
        \label{fig:all_association_hm_color_fruit}
    \end{subfigure}

    \centering
    \begin{subfigure}[b]{0.48\textwidth}
        \centering
        \includegraphics[width=\textwidth]{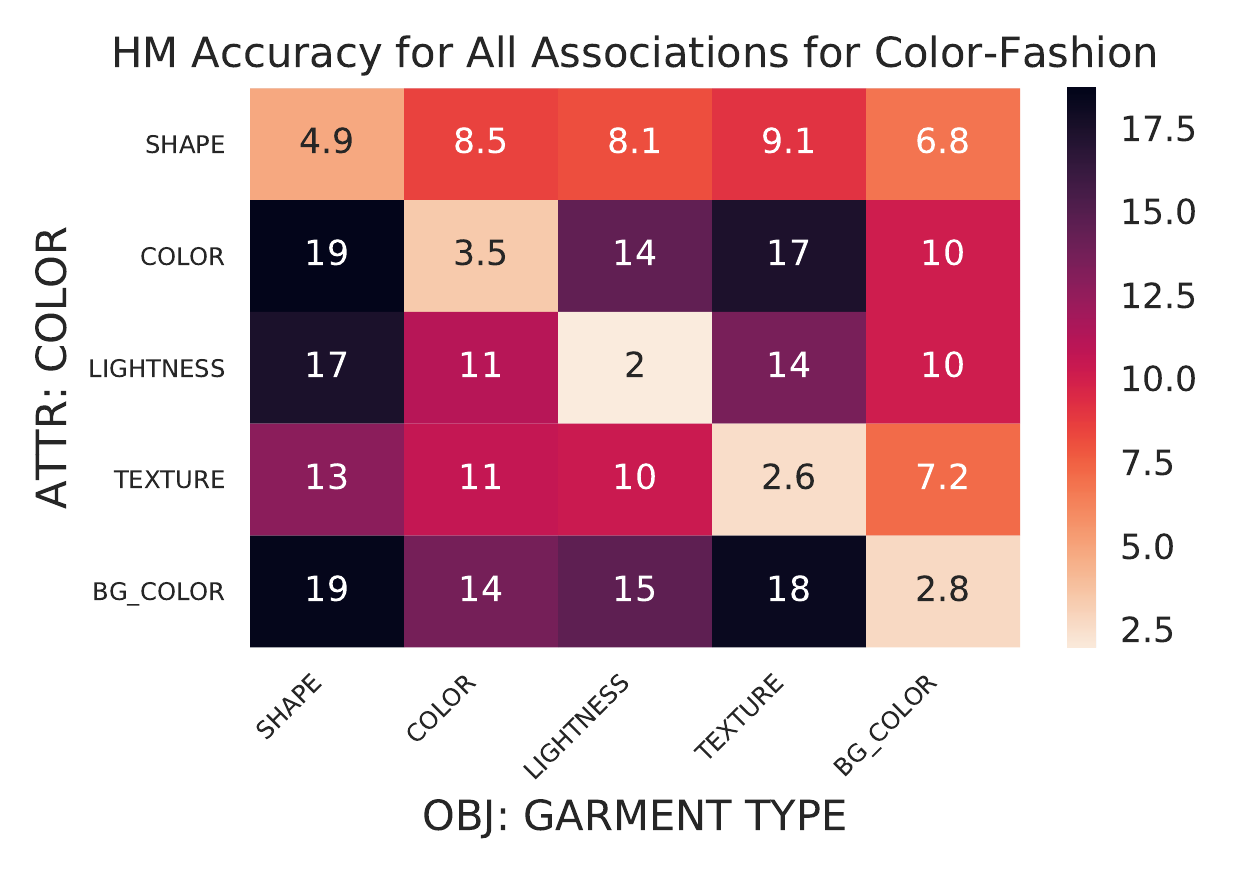}
        \caption{Color-Fashion}
        \label{fig:all_association_hm_color_fashion}
    \end{subfigure}
    
    \caption{HM Accuracies with different configurations of association matrices in different target datasets. The cell $C_{ij}$ in each heatmap corresponds to the accuracy when the association matrix associating target attribute and object type to the $i$-th and $j$-th factor respectively. The order of factors is shape, color, lightness, texture and background color.}
    \label{fig:all_association_hm}
\end{figure*}

We hypothesize that the shape is a generic robust factor that can be used to predict object types. So, we manually associate the shape factor from the source domain to the object type of target domains in the association matrix $A$ in all fully-correlated target domain scenarios. To validate this hypothesis, we perform an ablation study to evaluate network performance when different configurations of source factors are chosen in association matrix $A$. Performance of all configurations can be visualized as two-dimensional heatmaps for different datasets as in Figure \ref{fig:all_association_hm}. The value in each cell $C_{ij}$ of a heatmap represents the average HM accuracy when target attribute and target object associate to source factor $i$ (row of heatmap) and $j$ (column of heatmap) in the source domain respectively. In this regard, cell $C_{21}$ in each heatmap represents HM accuracy of a configuration setting when target attribute/object associate with source color/shape factors.

First of all, considering the results from DiagVib-Animal and Color-Fruit target datasets in Figure \ref{fig:all_association_hm_dv} and \ref{fig:all_association_hm_color_fruit}, the highest values of $C_{21}$ (42\% and 58\%) for both datasets empirically support our hypothesis that the shape factor is a robust factor for predicting object types in the target domain. In the Color-Fruit dataset, an interesting observation can be made as texture factor is also a predictive of fruit type in addition to the shape factor ($C_{13}$ of 40\% in Figure \ref{fig:all_association_hm_color_fruit}). This result shows capability of the texture to predict fruit type which aligns to the estimated association matrix in Figure \ref{fig:learn_association_matrix_color_fruit_la_r_1}. From these results, we can empirically validate our hypothesis and show that a proper configuration of the association matrix is important to alleviate model vulnerability to shortcuts.

In a more challenge dataset Color-Fashion, even though our configuration $C_{21}$ is among the best, there are other configuration settings that reach similar result (Figure \ref{fig:all_association_hm_color_fashion}). This behavior can be explained intuitively: considering the target object type (garment type), models have high performance when associating the object type to either shape or texture factors (can be seen as cells of high values on the first and the fourth columns). This behaviour is similar to the case of the fruit type in Color-Fruit dataset emphasizing the fact that both shape or texture can be a predictive factor for object types. For the target attribute type (garment color), its associations to color or background color produce high accuracies (can be seen as cells of high values on the second and the fifth rows). This implies that information of the garment color is contained in factor representations of both color and background color. The underlying reason can be due to the design of our source domain. In the DiagVib-Caltech source domain, boundaries between foregrounds and backgrounds are simple as backgrounds are only plain colors. However, in the case of the Color-Fashion target domain, its backgrounds are more complex representing realistic scenes. This suggests redesigning of the source domain. One possibility is to use more realistic backgrounds such as place images similar to \cite{ahmed2020systematic}.

\subsection{Implementation Details}
\label{sec:appendix_implementation}

In this section, we provide details of our network design and training hyperparameters.

For all variants of the factorized architecture illustrated in Figure \ref{fig:architecture_factorized} (Factor-0, FactorSRC, FactorSRC-CI and FactorSRC-IL), the encoder $G$ is a fully-connected network with 2 hidden layers, which outputs multiple factor representations, each one of length 64. All branches of $H_s$ and $H_t$ (i.e., all prediction heads in $\{{h}_s^{k}\}_{k=1}^K \cup \{{h}_s^{o}, {h}_s^{a}\}$) consist of a fully-connected network with 1 hidden layer. We set the hyperparameter $\lambda$ equal to 10 when we include the source dataset for all experiments for fair comparison.  In section \ref{sec:learning_factor_association}, we introduce strategy to learn the factor association matrix with additional regularization constraints. Hyperparameters $\alpha$, $\beta$ and $\tau$ used for the regularization constraints are $5$, $20$ and $0.33$ respectively.

For training we use Adam as an optimizer, a learning rate of 0.01 and weight decay equal to $5e^{-5}$. The optimal network is selected based on the loss on a validation split over 100 epochs.

\subsection{Bias Terms for Adjusting Likelihood of Unseen Combinations}
\label{sec:appendix_evaluation_with_bias}

\begin{table*}[t]
\caption{Accuracies on DiagVib-Animal, Color-Fruit, AO-CLEVR and Color-Fashion target domains with the similar experiment setup as in Table \ref{table:performance_overview}. However, calibrated bias terms are incorporated before computing seen, unseen and HM accuracies.}
\centering
\fontsize{7}{11}\selectfont
\begin{tabular}{cc|ccc|ccc|ccc}
\Xhline{4\arrayrulewidth}
\multirow{2}{*}{Approach} & \multirow{2}{*}{\makecell{Use \\ Source?}} & \multicolumn{3}{c|}{DiagVib-Animal} & \multicolumn{3}{c|}{Color-Fruit} &  \multicolumn{3}{c}{Color-Fashion} \\ 
 && Seen & Unseen & HM & Seen & Unseen & HM & Seen & Unseen & HM \\ \hline
LabelEmbed+ & \ding{55} & 96.3 & 10.3 & 13.2 & 100 & 19.7 & 12.5 & 90.0 & 13.4 & 16.9 \\
TMN & \ding{55} & 95.7 & 7.0 & 12.2 & 100 & 17.9 & 29.2 & 89.4 & 7.0 & 8.9 \\
CGE & \ding{55} & 92.8 & 11.8 & 15.0 & 100 & 24.9 & 32.9 & 88.1 & 20.8 & 21.0  \\ \Xhline{4\arrayrulewidth} 
\end{tabular}
\label{table:performance_overview_wb}
\end{table*}

\begin{figure*}[t]
    \centering
    \begin{subfigure}[b]{0.45\textwidth}
        \centering
        \includegraphics[width=\textwidth]{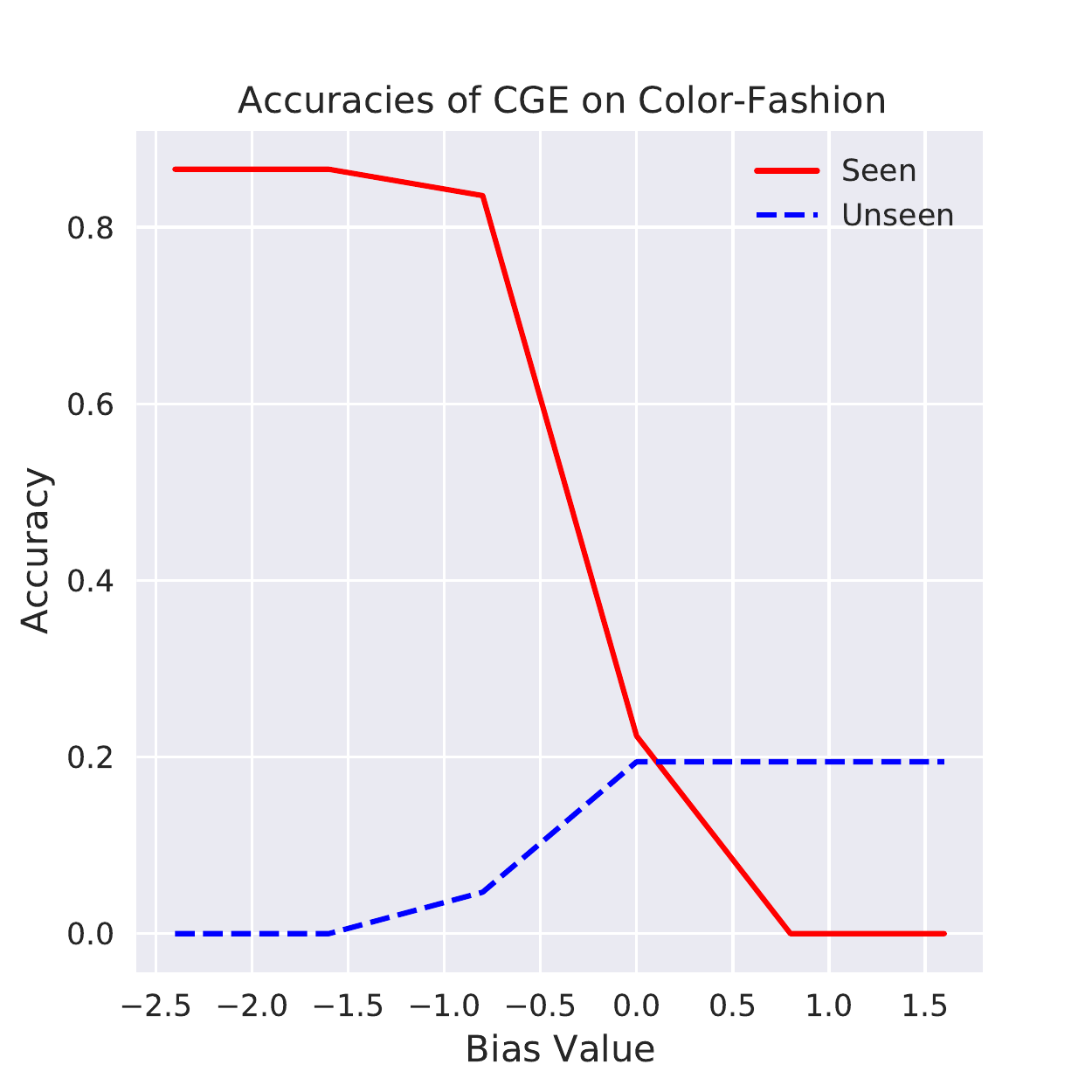}
        \caption{Seen/Unseen Accuracies of CGE}
        \label{fig:bias_sweep_cge}
    \end{subfigure}
    \begin{subfigure}[b]{0.45\textwidth}
        \centering
        \includegraphics[width=\textwidth]{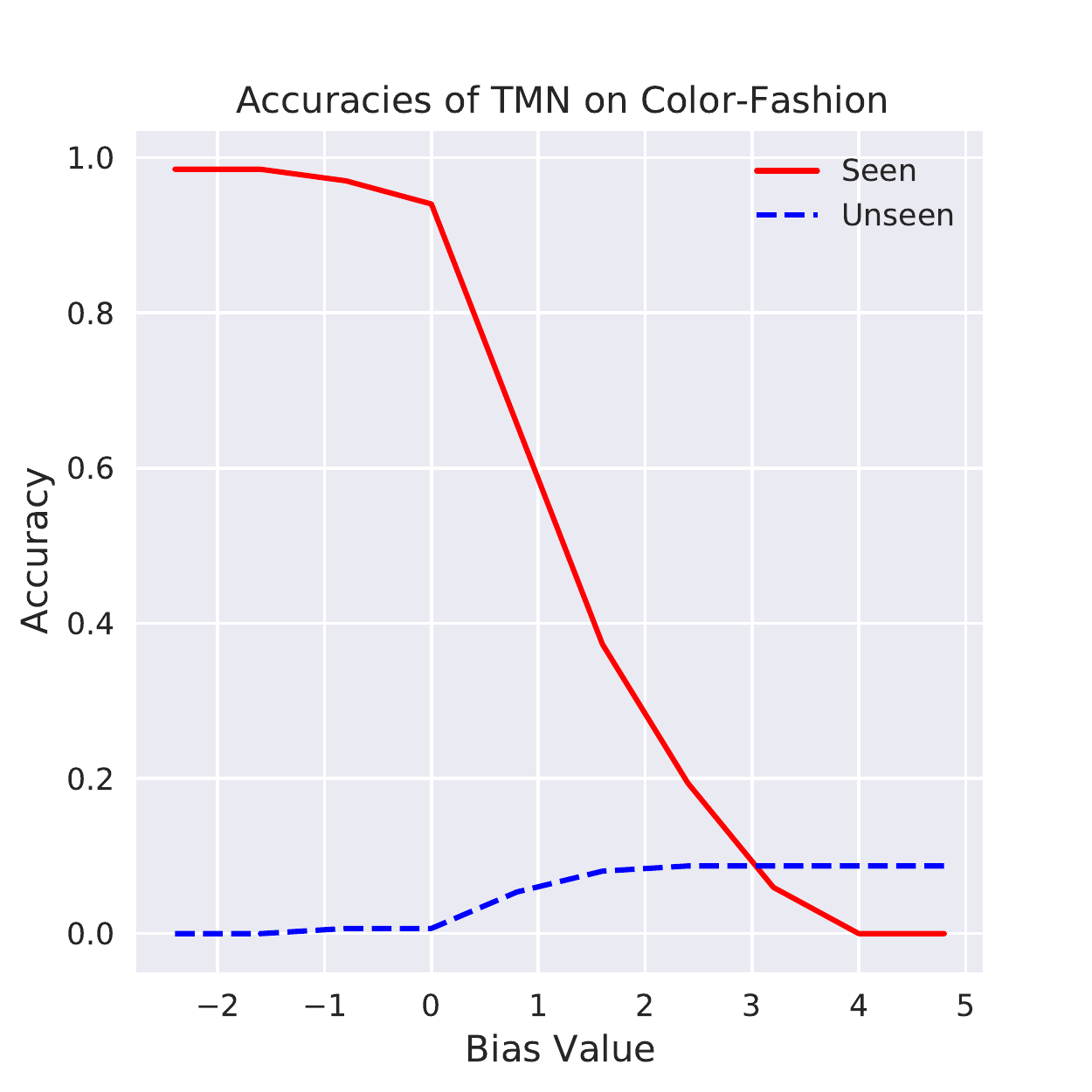}
        \caption{Seen/Unseen Accuracies of TMN}
        \label{fig:bias_sweep_tmn}
    \end{subfigure}    
    \caption{Seen/Unseen Accuracies of TMN and CGE baselines evaluated with different bias terms.}
    \label{fig:bias_sweep}
\end{figure*}

As mentioned earlier in section \ref{sec:experiments}, unlike some prior works \cite{purushwalkam2019task,naeem2021learning}, we evaluate compositional generalization without bias terms to adjust the likelihood of unseen combinations (using higher bias makes the model more likely to predict unseen combinations). The reason is that tuning of the bias terms requires availability of samples from unseen combinations. This violates the zero-shot assumption. Additionally, bias terms are designed to be applicable only with certain baselines based on compatability scores (such as LabelEmbed+, TMN and CGE) but not the others leading to unfair comparison.

For completeness, we will also provide results when the calibrated bias terms are incorporated during evaluation for LabelEmbed+, TMN and CGE. The seen, unseen and HM accuracies reported here correspond to their maximum values when their optimal bias terms are used (maximum seen and maximum unseen accuracies usually employ different optimal values of bias terms). Adopting the same experiment setup similar to Table \ref{table:performance_overview}, baseline performance with calibrated bias terms is presented in Table \ref{table:performance_overview_wb}. According to the results, the accuracies are higher when the calibrated biases are incorporated. However, the overall HM accuracies are still lower than results from our approaches. This still highlights vulnerability of these baselines to shortcuts.

Here, we also investigate why seen accuracies of certain baselines are low in Table \ref{table:performance_overview} (e.g., CGE on Color-Fashion). We can understand this behavior by observing seen/unseen accuracies using different bias terms. According to Figure \ref{fig:bias_sweep_cge}, the seen accuracy of CGE on Color-Fashion can be as high as 88.1 (similar to Table \ref{table:performance_overview_wb}) when low bias term is used. However, in our experiment, we choose not to use bias terms for evaluation as per the reasons described above. Therefore, the reported seen accuracies on Table \ref{table:performance_overview} are computed with bias terms of zero values. From Figure \ref{fig:bias_sweep_cge}, the seen accuracy of CGE on Color-Fashion is reduced to 21.6 (similar to Table \ref{table:performance_overview}). In contrast to CGE, the seen accuracy of TMN with zero bias term is already high (see Figure \ref{fig:bias_sweep_tmn}). Therefore, we do not see low seen accuracy of TMN on Table \ref{table:performance_overview}.


\subsection{Sweeping Weight of Loss for the Source Domain}
\label{sec:appendix_sweep_lambda}

\begin{figure}[!ht]
	\centering
    \includegraphics[width=0.45\textwidth]{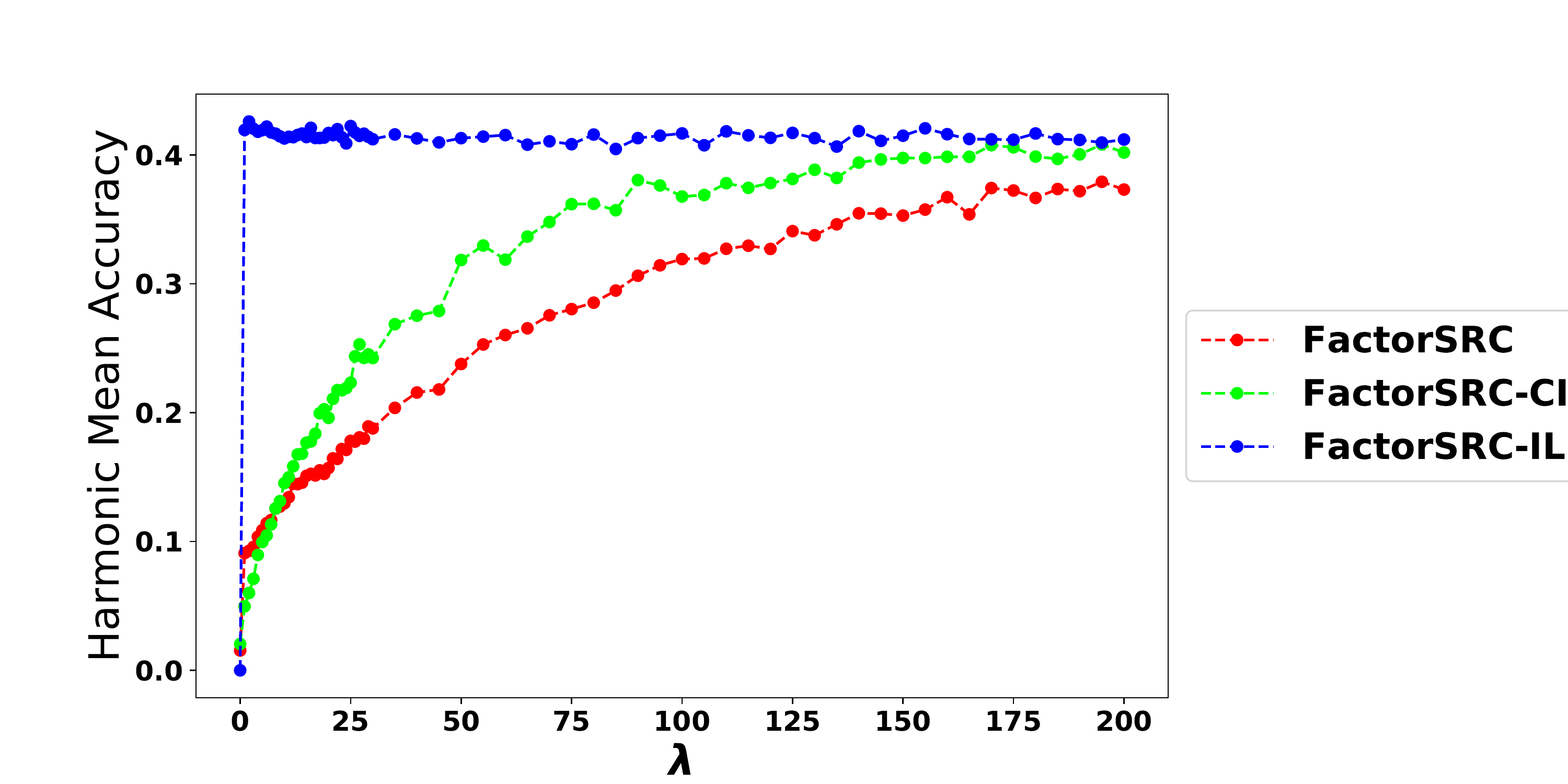}
    \caption{HM Accuracies using the DiagVib-Animal target domain and the DiagVib-Caltech source domain with different $\lambda$ values to weight the importance of $\mathcal{L}_{source}$. Notice that, higher $\lambda$ values encourage models to behave closer to FactorSRC-IL. We sample $\lambda$ on the low values with higher frequency to better highlight the faster increasing trends.}
    \label{fig:lambda_sweep}
\end{figure}

The hyperparameter $\lambda$ is used to weight the importance of $\mathcal{L}_{source}$ during training. Here we investigate its impact on the generalization performance attained in the target domain. Results of our analysis are shown in Figure \ref{fig:lambda_sweep}. We note that, for FactorSRC and FactorSRC-CI, the harmonic mean of seen and unseen accuracies increases with higher $\lambda$ values. This suggests that these two models are less sensitive to biases in the target dataset when $\lambda$ is increased. High values of $\lambda$ encourage FactorSRC and FactorSRC-CI to be more similar to FactorSRC-IL as $\mathcal{L}_{target}$ becomes less important to update $G$ relative to $\mathcal{L}_{source}$. FactorSRC-IL, on the other hand, performs consistently when $\lambda > 0$. This result is reasonable since, when the IL constraint is introduced, $\mathcal{L}_{source}$ and $\mathcal{L}_{target}$ are independently used to update different parts of the network (they update $\left\{ G, H_s\right\}$ and $\left\{H_t\right\}$ respectively). We note that, even though the higher $\lambda$ leads to better performance, we reserve to use $\lambda$ at 10 in our experiment so that we can study effects from other loss terms. It should be noted that changing the value of $\lambda$ here does not play a major role in our analysis since the key trends would be the same.

\subsection{Cross-Factor Independence Constraint Algorithm}
\label{sec:appendix_ci_algorithm}

The Cross-Factor Independence constraint is implemented as a two-step optimization approach. 
In the first step, we update $H'$ by minimizing the sum of cross entropy loss terms for all cross-factor predictions, i.e.,
\begin{align}
\mathcal{L}_{H'} = \sum_{\forall k_1, k_2; k_1 \neq k_2} CE(H'_{k_{1}k_{2}}(z_{k_1}), y_s^{k_2})\;.
\end{align}
Subsequently, we update the whole network by minimizing the combination of $\mathcal{L}_{target}$, $\mathcal{L}_{source}$, and an additional independence loss $\mathcal{L}_{CI}$. In principle, $\mathcal{L}_{CI}$ could be formulated as $-\mathcal{L}_{H'}$ but we found that this leads to training instabilities due to the fact that such a loss is unbound. Instead, we minimize the cross entropy between the predictions of $H'$ and a uniform label distribution. This encourages each factor representation to be uninformative with respect to all other factors. Mathematically, $\mathcal{L}_{CI}$ can be written as follows:
\begin{align}
\mathcal{L}_{CI} = \gamma \sum_{\forall k_1, k_2; k_1 \neq k_2} CE\left(H'_{k_{1}k_{2}}(z_{k_1}), \frac{\mathbf{1}^{N_{\mathcal{F}_s^{k_2}}}}{N_{\mathcal{F}_s^{k_2}}}\right)
\end{align}, where $\mathbf{1}^{N}$ indicates a vector of ones with length $N$, $N_{\mathcal{F}_s^{k}}$ is the number of factor values of the $k$-th factor and $\gamma \geq 0$ is a hyperparameter (we use $\gamma = 5$). 

\subsection{Seen Accuracy form FactorSRC-IL on DiagVib-Animal}

According to the result from Table \ref{table:performance_overview}, we notice that, on DiagVib-Animal, even though the HM accuracy of FactorSRC-IL is significantly higher than all other approaches, the seen accuracy is dropped significantly (to $56.3\%$). The drop of the seen accuracy only presents in the case of DiagVib-Animal but not other target domains. We suspect that this behavior could stem from the lower random chance accuracy ($1\%$ on DiagVib-Animal compared to $4\%$ and $4.7\%$ on other target domains) or just the complexity of the DiagVib-Animal (with high intra-class variations and various backgrounds). In this regard, we conduct an experiment with reduced number of attribute/object labels from 10 to 5 so that it has the random chance accuracy of $4\%$ which is the same as the one of Color-Fruit. In this regard, seen, unseen and HM accuracies on this reduced version of the DiagVib-Animal target domain are $74.2\%$, $52.6\%$ and $61.4\%$ respectively. Notice that, the seen accuracy is higher than the one on the original version but it is still relatively lower compared to the seen accuracies on other target domains. We can, therefore, conclude that the lower of the seen accuracy on DiagVib-Animal stems not only from its lower random chance accuracy but also from the complexity of the target domain itself.

\subsection{Color-Fruit Dataset Generation}
\label{sec:appendix_fruit_generation}

\begin{figure*}[t]
    \centering
    \begin{subfigure}[b]{0.45\textwidth}
        \centering
        \captionsetup{justification=centering}
        \includegraphics[width=\textwidth]{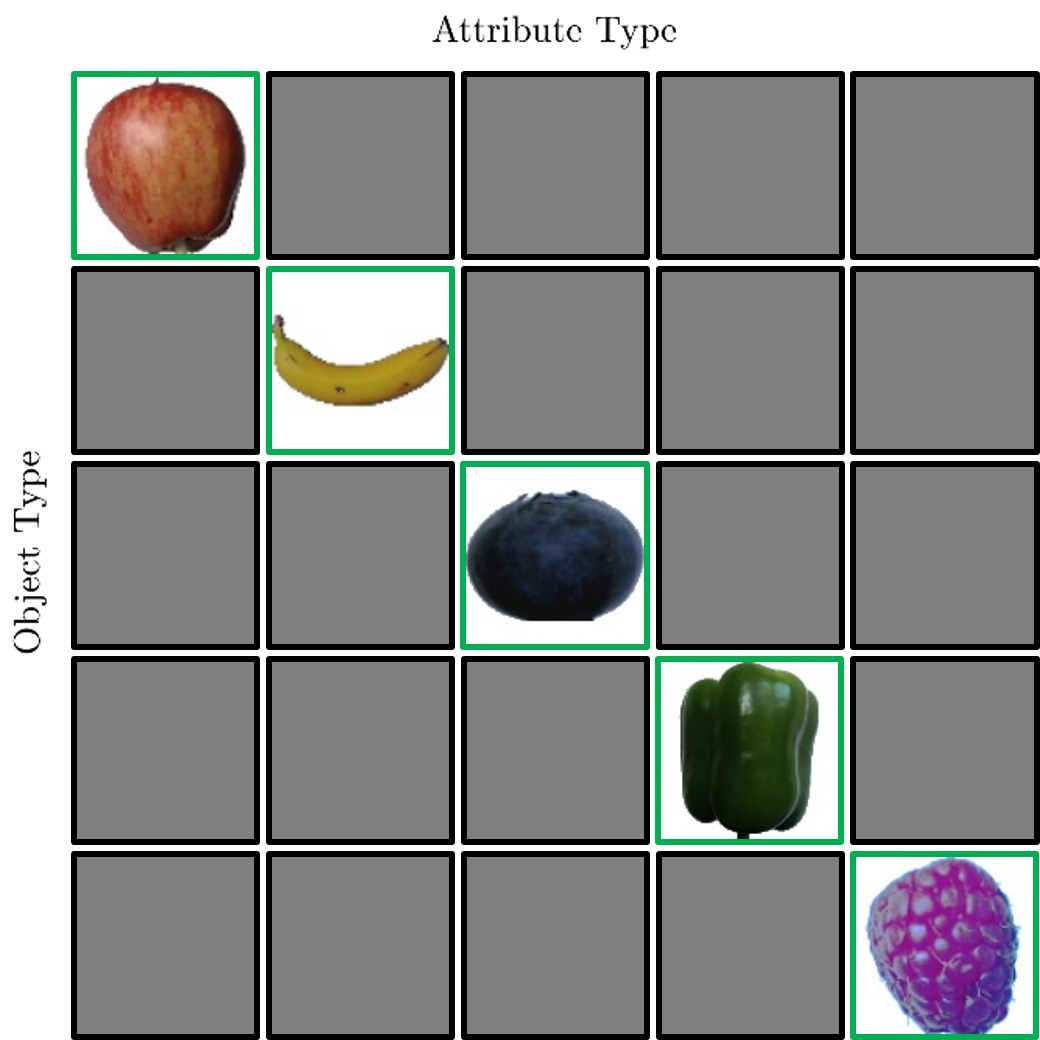}
        \caption{Training samples}
        \label{fig:fruit_correlation_details_train}
    \end{subfigure}
    \begin{subfigure}[b]{0.45\textwidth}
        \centering
        \captionsetup{justification=centering}
        \includegraphics[width=\textwidth]{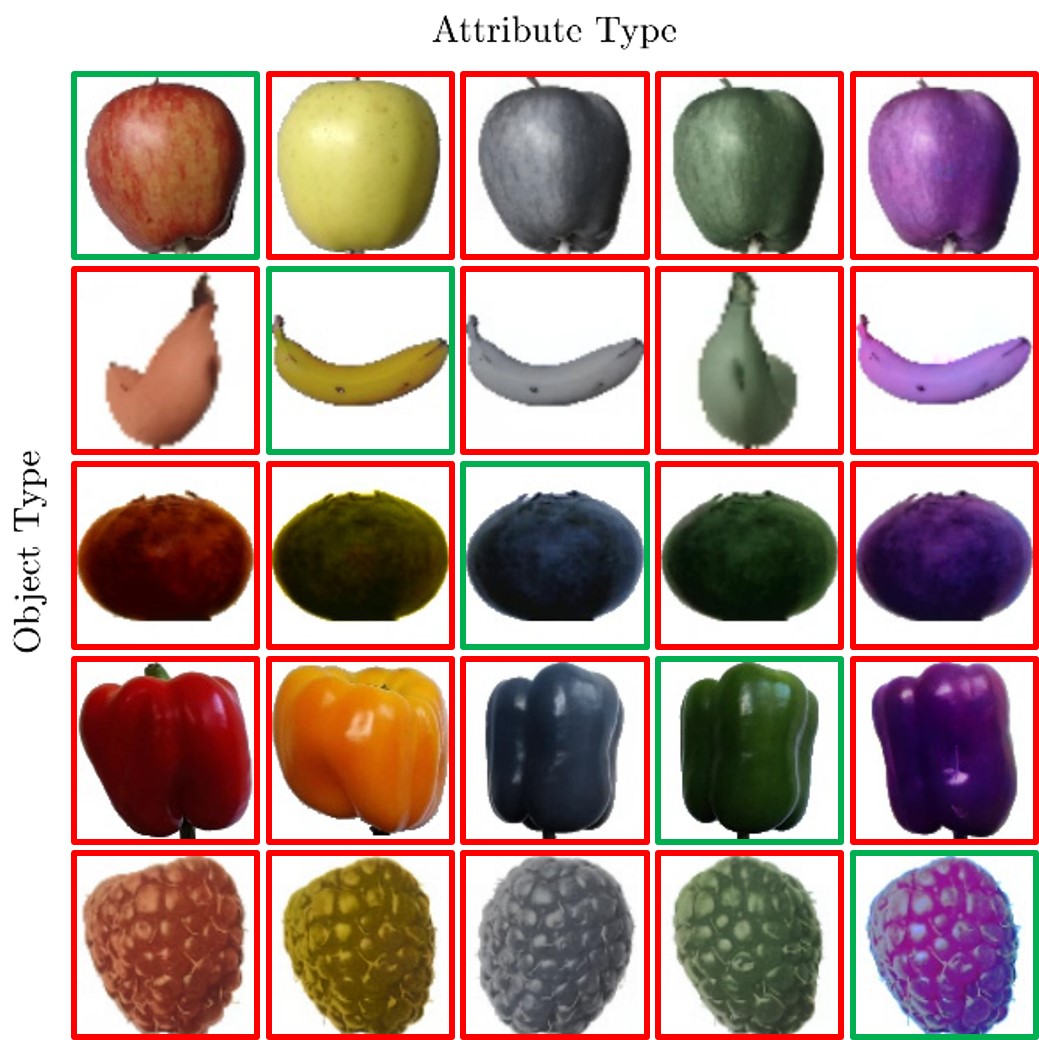}
        \caption{Test samples}
        \label{fig:fruit_correlation_details_test}
    \end{subfigure}
    \caption{Examples of Color-Fruit dataset images. (a) Training samples containing images of fully-correlated attribute-object combinations denoted with green-bordered boxes (One object type always has one color and vice versa). (b) Test samples are, on the other hand, uncorrelated i.e., consisting of images with any attribute-object combinations (i.e., each object (fruit type) can appear with any attributes (color)). Fruit images whose colors are not available in the original Fruits 360 dataset are obtained by using the recolorization technique in \cite{zhang2017real}}
    \label{fig:fruit_correlation_details}
\end{figure*}

In order to generate the \emph{Color-Fruit} dataset, used in our experiments, we use fruit images from the Fruits 360 dataset \cite{murecsan2017fruit}. Five fruits (Apple, Banana, Blueberry, Pepper and Raspberry) are selected as they have distinct colors (red, yellow, blue, green and magenta), which facilitate the evaluation of compositional generalization in the case of fully-correlated seen combinations. 

During evaluation, however, fruits with different colors are required. Thus, we perform recolorization of images in the test split using the approach described in \cite{zhang2017real}. Basically, an original test image is recolorized into median colors of all other fruits (e.g. a banana image is transformed such that it has a color similar to that of an apple, a blueberry, a pepper and a raspberry). More detailed visualization of the dataset is presented in Figure \ref{fig:fruit_correlation_details}.

%
%

\end{document}